%% file: main.tex
\documentclass[runningheads]{llncs}

\usepackage{eccv}

\usepackage{eccvabbrv}

\usepackage{graphicx}
\usepackage{booktabs}
\usepackage{multirow}
\usepackage{utfsym}
\usepackage[ruled,vlined]{algorithm2e}
\usepackage{colortbl}

\usepackage[accsupp]{axessibility}  

\usepackage{hyperref}

\usepackage{orcidlink}

\begin{document}

\title{MVPGS: Excavating Multi-view Priors for Gaussian Splatting from Sparse Input Views} 

\titlerunning{MVPGS}

\author{Wangze Xu\inst{1} \and Huachen Gao\inst{1} \and Shihe Shen\inst{1} \and Rui Peng\inst{1} \and Jianbo Jiao\inst{2} \and Ronggang Wang\inst{1,3}}

\authorrunning{W. Xu et al.}

\institute{School of Electronic and Computer Engineering, Peking University \and School of Computer Science, University of Birmingham \and Peng Cheng Laboratory \\
\email{xuwangze@stu.pku.edu.cn} \quad \email{rgwang@pkusz.edu.cn}}

\maketitle

\begin{abstract}
\input{contents/abstract}
\end{abstract}

\begin{figure}[htbp]
    \centering
    \includegraphics[width=120mm]{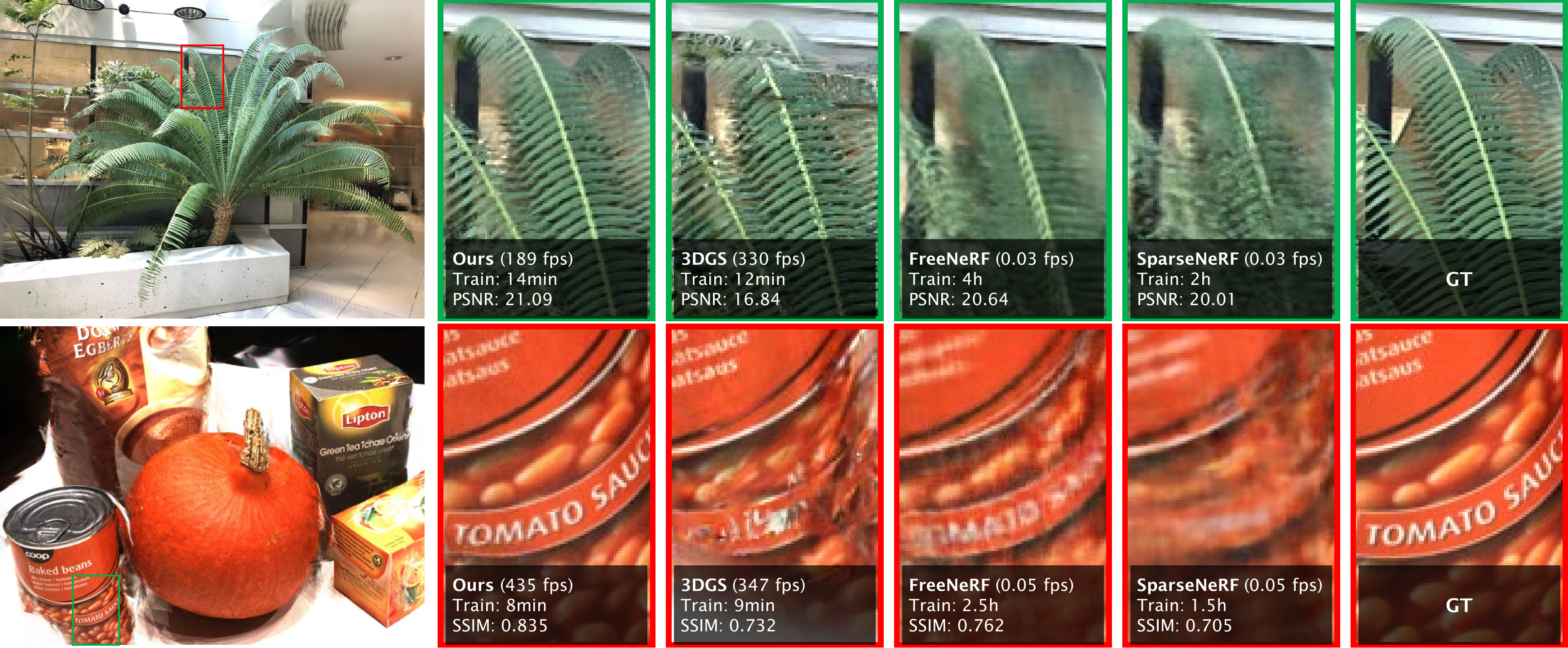}
    \caption{\textbf{Qualitative Results on LLFF and DTU Datasets under High-Resolution Setting with 3-view Inputs.} Compared with NeRF, the proposed method maintains high-fidelity quality in high-frequency regions and meanwhile achieves competitive training and rendering speed for few-shot NVS.}
    \label{fig:high_res_details}
\end{figure}

\section{Introduction}
\label{sec:intro}
\input{contents/introduction}

\section{Related Work}
\label{sec:related_work}
\input{contents/related_work}

\section{Method}
\label{sec:method}
\input{contents/method}

\section{Experiments}
\label{sec:exp}
\input{contents/experiment}

\section{Conclusion}
\label{sec:conclusion}
\input{contents/conclusion}

\par\vfill\par

%
%
\clearpage
\input{contents/acknowledgments}

\bibliographystyle{splncs04}
\bibliography{main}

\clearpage
\appendix
\section*{Supplementary Material}
\label{sec:supplementary}
\input{contents/supplementary}

\end{document}

%% file: contents/abstract.tex
Recently, the Neural Radiance Field (NeRF) advancement has facilitated few-shot Novel View Synthesis (NVS), which is a significant challenge in 3D vision applications. Despite numerous attempts to reduce the dense input requirement in NeRF, it still suffers from time-consumed training and rendering processes. More recently, 3D Gaussian Splatting (3DGS) achieves real-time high-quality rendering with an explicit point-based representation. However, similar to NeRF, it tends to overfit the train views for lack of constraints. In this paper, we propose \textbf{MVPGS}, a few-shot NVS method that excavates the \textbf{\underline{m}}ulti-\textbf{\underline{v}}iew \textbf{\underline{p}}riors based on 3D \textbf{\underline{G}}aussian \textbf{\underline{S}}platting. We leverage the recent learning-based Multi-view Stereo (MVS) to enhance the quality of geometric initialization for 3DGS. To mitigate overfitting, we propose a forward-warping method for additional appearance constraints conforming to scenes based on the computed geometry. Furthermore, we introduce a view-consistent geometry constraint for Gaussian parameters to facilitate proper optimization convergence and utilize a monocular depth regularization as compensation. Experiments show that the proposed method achieves state-of-the-art performance with real-time rendering speed. Project page: \url{https://zezeaaa.github.io/projects/MVPGS/}

%% file: contents/introduction.tex
Given a set of posed images, novel view synthesis(NVS) aims to render images of unseen views, which has shown great importance in 3D vision applications. Recently, the neural radiance field(NeRF)\cite{ori_nerf} brought large advancements to the NVS field, which implicitly represents a scene through MLPs and applies volume rendering to generate target views. Various subsequent methods have explored enhancements to NeRF\cite{mip_nerf,instant-ngp,mipnerf360}, yielding remarkable progress in practice. More recently, 3D Gaussian Splatting (3DGS)\cite{3dgs} employs a set of point-based Gaussians as scene representations, enabling efficient differentiable splatting for optimization and achieving high-quality, real-time rendering. While NeRF and 3DGS show impressive results in NVS, both methods require a substantial number of images for training. However, in real-world applications such as robotics and VR/AR, available views are often much sparser. In few-shot situations, these methods struggle to learn accurate 3D structures without enough constraints across views and the performance degrades drastically due to overfitting on training views\cite{dsnerf}. To achieve higher quality and efficiency in few-shot NVS, numerous methods have been developed and most of them are built upon NeRF for improvements. Some methods propose additional constraints\cite{infonerf,dietnerf,regnerf,geconerf} as regularization to prevent overfitting. Some introduce priors\cite{dsnerf,ddpnerf,scade,vipnerf,sparsenerf,darf}, such as depth, to guide NeRF to learn proper scene geometry. However, the rendering quality of these methods remains to be improved and they still suffer from time-consuming training and rendering processes due to the implicit representation and volume rendering.

In this paper, we propose MVPGS, a few-shot NVS method based on 3D Gaussian Splatting\cite{3dgs}, which leverages recent learning-based Multi-view Stereo\cite{mvsnet,casmvsnet,cvpmvsnet} to excavate more cues from sparse input views, achieving both high-quality and high-speed rendering. Similar to NeRF, 3DGS encounters overfitting problems during optimization under limited view constraints. However, unlike NeRF's blank MLPs, 3DGS's initial points contain part of the appearance and geometry priors of scenes. The proposed method explores the potential information from this explicit initialization to alleviate overfitting and preserve accurate geometries of scenes during optimization.

Specifically, we use MVSFormer\cite{mvsformer} to estimate dense view-consistent depths as initialization for 3D Gaussians. Our experiments (see in Tab. \ref{tab:all_quantitative_result}) show that the optimization of 3DGS benefits well from such richer initial information compared with original COLMAP's\cite{colmap1,colmap2} outputs, especially in few-shot situations. Furthermore, to mitigate overfitting, we utilize a forward-warping method\cite{FWD} to conduct additional appearance constraints conforming to scenes for other unseen observations, which can be seamlessly integrated with the initial positions of 3DGS derived from MVS. Since the warped locations might be floating-point numbers and not exactly aligned with an image grid, we utilize reversed bilinear sampling\cite{forward_backward_warp_optical_flow} to distribute colors to the local regions. Our approach differs from previous methods\cite{geoaugnerf,geconerf} (as illustrated in Fig.\ref{fig:fwd_warp}) where the appearance for constraints is fetched using the rendered depth of target view, which is called 'backward warping' as discussed in \cite{forward_backward_warp_optical_flow}. While these methods enhance the consistency between views, converging for large variations of camera poses in sparse views is difficult. Besides, compared to approaches\cite{regnerf,diffusionerf,deceptivenerf} that rely on plausible signals from other models, e.g. normalizing flow model\cite{flowmodel} or diffusion model\cite{diffusion}, to supervise the rendering of unseen views, our method provides appearance information that conforms to scenes through geometric solutions. More importantly, our approach naturally adapts to 3DGS's initialization and broadens the scope of available training data, even if it is derived from known views.

Furthermore, the position parameters in 3DGS are directly updated by the back-propagation gradient, which may lead to deviations from accurate geometry during few-shot optimization (see leaves in Fig. \ref{fig:ablation_visualization}). To facilitate convergence during optimization, we introduce a loss between 3DGS's geometry and the confident geometric structure computed from MVS. Additionally, MVS may have poor performance in areas such as textureless and low overlap\cite{mvsnet}, we incorporate monocular depth priors\cite{DPT} to further constrain the global geometry of scenes and mitigate the influence of inaccurate warped appearance priors caused by imprecise MVS depths.

Our contributions include
\begin{itemize}
    \item We propose a novel paradigm for few-shot Novel View Synthesis (NVS) which utilizes an efficient point-based Gaussian representation and is seamlessly integrated with geometric priors from Multi-View Stereo (MVS).
    \item We introduce forward warping for Gaussian initialization obtained from MVS to extract potential appearance information as constraints for unseen views. Additionally, we employ geometry constraints to ensure the proper convergence of 3D Gaussians.
    \item Experiments on \textit{LLFF}, \textit{DTU}, \textit{NVS-RGBD}, and \textit{Tanks and Temples} datasets demonstrate our method's state-of-the-art performance while maintaining real-time rendering speed.
\end{itemize}

%% file: contents/related_work.tex
\noindent\textbf{Neural Representations for 3D Reconstruction.} Neural Radience Field (NeRF)\cite{ori_nerf} has shown great power in novel view synthesis. Different from previous explicit scene representations such as point clouds, voxels, and meshes, NeRF is a compact paradigm that represents scenes with MLPs implicitly. Let a NeRF be a function $f$ that maps 3D coordinates $(x,y,z)$ and view direction $d$ to corresponding color $c$ and density $\sigma$. With the volume rendering formulation $C = \sum_{i}T_{i}\sigma_{i}c_{i}$\cite{ori_nerf}, we can derive each pixel's color by integrating colors and densities along a ray from the camera to the scene. NeRF brought a new paradigm for scene representations and tremendous follow-ups emerged focusing on improving the rendering quality\cite{mip_nerf,mipnerf360}, accelerating\cite{instant-ngp,dvgo}, 3D generation\cite{zero1to3,dream3d,consistent1to3}, dynamic scenes\cite{dnerf,kplanes}, and generalizing to unseen scenes\cite{srf,pixelnerf,mvsnerf,peng2023gens}. Among them, Mip-NeRF\cite{mip_nerf} replaces point-based ray tracing with cone tracing to alleviate aliasing and Mip-NeRF360\cite{mipnerf360} then extends Mip-NeRF to unbounded scenes. Though these methods achieve excellent rendering results, the training and rendering time (several days to hours) is unbearable. For accelerating, Instant-NGP\cite{instant-ngp} proposed a multi-resolution hash encoding to compress the training time to seconds. Recently, 3D Gaussian Splatting (3DGS)\cite{3dgs} has shown a great prospect for real-time rendering. Instead of time-consuming volume rendering, 3DGS takes point clouds as initialization and utilizes an alpha blending method for the rasterization. Benefitting from this explicit representation, 3DGS is highly efficient compared to NeRF. However, its performance degrades drastically with insufficient inputs due to the similar overfitting in NeRF. 

\noindent\textbf{Few-shot Novel View Synthesis.} There are numerous pieces of research to improve the performance of NeRF with sparse input views. Some methods add additional constraints to alleviate over-fitting. RegNeRF\cite{regnerf} applies appearance regularization from normalizing flow model\cite{flowmodel} and conducts depth smooth loss on unseen views. FreeNeRF\cite{freenerf} proposes a frequency regularization to solve the overfitting problem caused by high-frequency components. ViPNeRF\cite{vipnerf} revises the framework of the original NeRF to output the visibility of points and regularize visibilities and depths on input views. Another branch introduces priors such as depths or semantics to supervise NeRF to learn proper information about scenes\cite{ddpnerf,scade}. DS-NeRF\cite{dsnerf} firstly uses sparse depth generated from COLMAP\cite{colmap1,colmap2} to supervise the depth distribution along a ray. SparseNeRF\cite{sparsenerf} utilizes an effective scale-invariant loss to distill geometric cues from monocular depth estimation. Diet-NeRF\cite{dietnerf} uses CLIP\cite{clip} to add a semantic consistency between views. Different from previous few-shot NVS methods, our method is built on explicit point-based representation and achieves both real-time rendering and competitive rendering quality.

\noindent\textbf{Muti-View Stereo Reconstruction.} Recent learning-based MVS\cite{mvsnet, casmvsnet, cvpmvsnet, peng2022rethinking, zhang2023geomvsnet, mvsformer, xiong2023cl} has demonstrated remarkable advancements in 3D reconstruction. Typically, it constructs a cost volume by warping 2D image features into different front-parallel planes of the reference camera. Subsequently, 3D CNNs are employed for cost regularization and depth regression\cite{mvsnet}. MVS exhibits better robustness to noise and occlusions, achieving higher accuracy compared to traditional methods\cite{colmap1,colmap2} that rely on handcrafted features for matching. Our approach leverages the MVS method\cite{mvsformer} enhanced with Vision Transformers (ViTs) to thoroughly excavate clues from few-shot inputs and facilitate the optimization of 3DGS.

%% file: contents/method.tex
\begin{figure}[h]
    \centering
    \includegraphics[width=120mm]{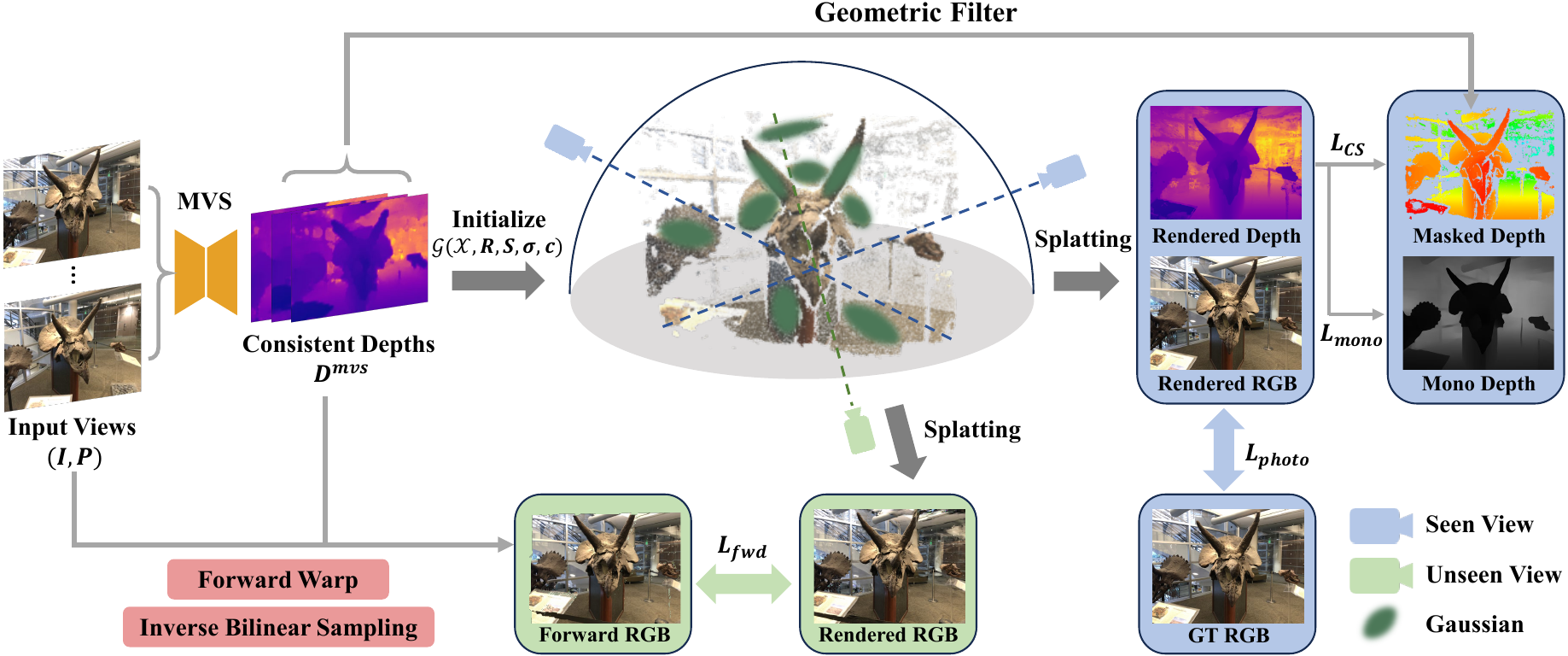}
    \caption{\textbf{Framework Overview.} MVPGS leverages learning-based MVS to estimate dense view-consistent depth $D^{mvs}$ and construct a point cloud $\mathcal{P}$ for the initialization of Gaussians $\mathcal{G}$. We excavate the computed geometry from MVS through forward warping to generate appearance priors for the supervision of unseen views. To regularize the geometry update during optimization, we introduce $L_{CS}$ from MVS depth and $L_{mono}$ from monocular depth priors to guide Gaussians to converge to proper positions.}
    \label{fig:framework}
\end{figure}

Fig. \ref{fig:framework} illustrates the overall framework of MVPGS. Our approach is built upon the highly efficient 3D Gaussian splatting\cite{3dgs}, which employs point-based explicit representation for scenes (Sec. \ref{sub_sec:gaussian_splatting_preliminary}). To address overfitting in few-shot scenarios, we utilize a forward warping method to extract appearance information as constraints for unseen views based on geometry computed from MVS (Sec. \ref{sub_sec:multi_view_priors}). Moreover, we conduct view-consistent geometric regularization to preserve the accurate structure of Gaussian parameters and apply monocular depth priors as compensation during optimization (Sec. \ref{sub_sec:geometric_reg}).

\subsection{Preliminary for Gaussian Splatting}
\label{sub_sec:gaussian_splatting_preliminary}
3D Gaussian Splatting\cite{3dgs} represents scenes explicitly with a set of point-based Gaussian parameters. Each Gaussian distribution is defined by a 3D covariance matrix $\Sigma$ and a center position at point (mean)  $\mathcal{X}$ in world space\cite{3dgs}:
\begin{equation}
    G(X)=e^{-\frac{1}{2}\mathcal{X}^T\Sigma^{-1}\mathcal{X}}.
\end{equation}
To reduce the difficulty of optimizing, $\Sigma$ is decomposed into two learnable parameters with physical significance:
\begin{equation}
    \Sigma = \mathbf{R}\mathbf{S}\mathbf{S}^T\mathbf{R}^T,
\end{equation}
where $\mathbf{R}$ is a quaternion rotation matrix and $\mathbf{S}$ is a scaling matrix. Besides, each Gaussian also has attributions of opacity $\sigma$ and radiance $c$ represented by spherical harmonics (SH) for rendering. The complete Gaussian parameter is defined by $\mathcal{G}=\{(\mathcal{X}_{i},\mathbf{S}_{i},\mathbf{R}_{i},\sigma_{i},c_{i})\}_{i=0}^{i=n}$.

When rendering, a splatting technique\cite{3dgs} is utilized and each 3D Gaussian is projected to 2D image space for rasterization. With a viewing transform $W$, we can obtain the covariance $\Sigma'$ in camera coordinate:
\begin{equation}
    \Sigma^{\prime} = JW\Sigma W^TJ^T,
\end{equation}
where $J$ is the Jacobian of the affine approximation of the projective transformation. After projecting Gaussians into 2D, all Gaussians covering a pixel are sorted and each pixel's color $C$ is computed by blending the overlapping Gaussians:
\begin{equation}
    \label{eq:3dgs_color}
    C = \sum_{i\in N} T_i \alpha_i c_i,
\end{equation}
where $T_i = \prod_{j=1}^{j=i-1} (1-\alpha_j)$ is the transmittance, $\alpha_i$ is $\sigma_{i}$ muliplied by a 2D Gaussian with covariance $\Sigma^{\prime}$, and $c_i$ is $i-th$ Gaussian's color based on spherical harmonics. The photometric loss for optimizing parameters is given by a combination of $L_{1}$ and SSIM\cite{ssim} loss:
\begin{equation}
    \label{eq:ori_img_loss}
    L_{photo} = \lambda_{1} L_{1}(\hat{I},I) + (1-\lambda_{1})(1-SSIM(\hat{I},I))
\end{equation}
where $\hat{I}$ and $I$ are the rendered image and the ground-truth, respectively. $\lambda_{1}$ is a controlling weight which is set to 0.8 in experiments.

\subsection{Multi-view Stereo Priors for Optimization}
\label{sub_sec:multi_view_priors}
\subsubsection{3D Gaussian Initialization.}
\label{subsub_sec:gaussian_init}
We leverage MVSFormer\cite{mvsformer}, a learning-based MVS method enhanced by Vision Transformers (ViTs), to excavate more cues from sparse inputs. Given a set of input views $\{V_{i}^{train}=(I_{i}, P_{i})\}_{i=0}^{i=N}$ (where $I,P$ denote images and corresponding camera pose, respectively), we consider each view $V_{i}$ as reference view and the remaining as source views. For each reference view, we warp all source views' 2D features to construct a 3D cost volume and regress the depth map $D_{i}^{mvs}$. Then we employ geometric filter\cite{mvsnet} based on depth reproject errors and areas with high consistency are indicated by masks $\{M_{i}\}_{i=0}^{i=N}$. With $\{D_{i}^{mvs}\}$ and $\{M_{i}\}$, we use method in \cite{mvsnet} to merge them into a set of 3D positions $\{(x_i,y_i,z_i)\}_{i=0}^{i=n}$, which is regarded as the mean of initial 3D Gaussian's $\mathcal{X}$. More details are in the supplementary material (Supp. Mat.).

\subsubsection{Forward Warping Appearance for Unseen Views.}
\label{sub_sec:useen_priors}
In contrast to NeRF's lack of scene knowledge, the initialization of Gaussian parameters incorporates potential priors about scenes. Through the ViT-based MVS model, we can acquire richer scene information compared to the original COLMAP outputs, particularly in sparse input scenarios. We further exploit these priors to alleviate overfitting in few-shot situations. The key idea is to leverage the computed geometry from MVS to obtain the appearance of unseen views around training views by forward warping, which can be seamlessly integrated into 3DGS's pipeline. Our approach differs from previous methods\cite{geoaugnerf,geconerf}, which is termed 'backward warping' as discussed in \cite{forward_backward_warp_optical_flow} and is used widely in self-supervised monocular depth estimation\cite{zhou2017unsupervised}. Backward warping utilizes the target view's geometry to collect appearance from other views, while the proposed method infers the target view's appearance based on the known view's geometry and appearance.

\begin{figure}[t]
    \centering
    \includegraphics[width=120mm]{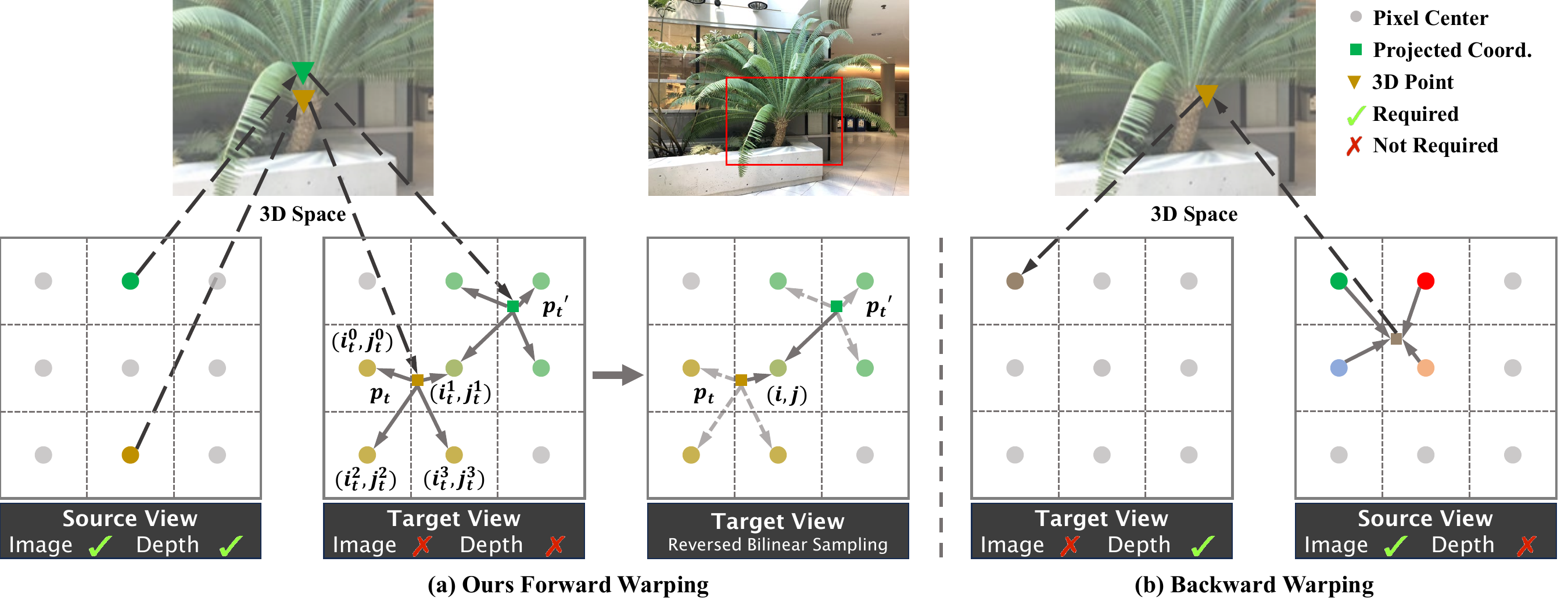}
    \caption{\textbf{Forward Warping and Backward Warping.} Forward warping (a) takes $(I_{src},D_{src}^{mvs},P_{src},P_{tgt})$ as input and output the appearance $I_{tgt}$ in pose $P_{tgt}$ based on the known geometry $D_{src}^{mvs}$. Owning to the warped location $p_{t}$ is not religiously lie in the center of the grid, we adopt reversed bilinear sampling\cite{forward_backward_warp_optical_flow} to determine the color of each pixel. This is different from backward warping (b) which uses bilinear sampling to sample color from the source view according to the target view's geometry.}
    \label{fig:fwd_warp}
\end{figure}

Given a known training view $V_{src}$ with corresponding image $I_{src}$, MVS depth $D_{src}^{mvs}$, camera pose $P_{src}$ and target view's pose $P_{tgt}$, the forward warping is defined as
\begin{equation}
    \label{eq:fwd}
    I_{tgt}=fwd(I_{src},D_{src}^{mvs},P_{src},P_{tgt}).
\end{equation}
where $I_{tgt}$ is the image in target view $V_{tgt}$ warped from $V_{src}$. To establish a more robust mapping relationship between pixels in two views, we combine forward warping with reversed bilinear sampling\cite{forward_backward_warp_optical_flow}. The process is concluded as follows: 1) find the corresponding locations in the target view for each pixel in $V_{src}$, 2) find the nearest pixels influenced by the warped location, 3) compute the color contribution to the target view's pixels\cite{kanchana2022revealing}. Specifically, the coordinate relationships between $V_{src}$ and $V_{tgt}$ is
\begin{equation}
    \label{eq:homogeneous}
    p_{tgt} \sim KP_{tgt}^{-1}P_{src}D_{src}^{mvs}K^{-1}p_{src}.
\end{equation}
where $K$ is the camera intrinsic. $p_{src}$ and $p_{tgt}$ denote the pixel coordinates in $V_{src}$ and the warped coordinates in $V_{tgt}$. Given a pixel $p_{s}=(x_{s},y_{s}) \in p_{src}$ and its depth $d_{s}^{mvs}$ in $V_{src}$, $p_{s}$ can be backproject to a 3D point $\tau$ and then projected to $V_{tgt}$ to get the coordinate $p_{t}=(x_{t},y_{t})$. The color $c_{t}$ at position $p_{t}$ can be viewed as the corresponding color $I_{src}(p_{s})$ in $V_{src}$, under the assumption that the color remains constant across different views. In practice, $p_{t}$ is fractional and may not religiously lie in the pixel center. To address this problem, we utilize reversed bilinear sampling\cite{forward_backward_warp_optical_flow}, which distributes the weight of the projected pixel to its nearest neighbors. Specifically, we choose the 4 nearest pixel centers $\{(i_{t}^{m},j_{t}^{m})\}_{m=0}^{m=3}$ around $p_{t}$ as the areas influenced by $p_{t}$ (illustrated in Fig. \ref{fig:fwd_warp}). The color's contribution to these 4 pixels from the warped point is determined by the distance\cite{forward_backward_warp_optical_flow}:
\begin{equation}
    \label{eq:reversed_bilinear_sampling}
    w_{t}(i,j) = (1-|i-x_{t}|)(1-|j-y_{t}|)
\end{equation}
With Eq. \ref{eq:reversed_bilinear_sampling}, pixels in target views can be colored with corresponding projected points $p_{t}$. Another problem is that different points may map to the same location which causes occlusion. To solve this, we align higher weights for points that are closer to the screen. Following \cite{kanchana2022revealing}, the depth weight is computed as $w_{t}^{d}=\frac{1}{(1+d_{t})^{\gamma}}$,
where $d_{t}$ is $p_{t}$'s depth in $V_{tgt}$. $\gamma$ is a hyperparameter given by $\gamma=\frac{50}{\log(1+d^{max})}$, where $d^{max}$ is the maximum of projected depths in $V_{tgt}$ for all $p_{t} \in p_{tgt}$\cite{kanchana2022revealing}. The final color of the target view pixel is 
\begin{equation}
    \label{eq:warped_pixel_color}
    I_{tgt}(i,j)=\frac{\sum_{p_{t} \in \Omega}w_{t}(i,j)w_{t}^{d}c_{t}}{\sum_{p_{t} \in \Omega}w_{t}(i,j)w_{t}^{d}},
\end{equation}
where $\Omega$ denotes all mapped points that contributes to pixel $(i,j)$. 

For additional supervision in unseen views, we follow \cite{regnerf} to sample random pose $P'$ and generate its forward warping appearance $I'$ according to the Eq. \ref{eq:warped_pixel_color}. During optimization, we render the image $\hat{I'}$ with $P'$ from Gaussian parameters. We use a combination of $L_{1}$ and SSIM loss for supervision:
\begin{equation}
    \label{eq:fowward_warp_img_loss}
    L_{fwd} = \lambda_{2} L_{1}(\hat{I'},I') + (1-\lambda_{2})(1-SSIM(\hat{I'},I'))
\end{equation}
where $\lambda_{2}$ is set to 0.2 in our experiments.

\subsection{Geometric Regularization}
\label{sub_sec:geometric_reg}
The geometry of scenes can be reflected by the Gaussian's mean parameter $\mathcal{X}$, which is updated directly through back-propagate gradients during optimization. In practice, these parameters tend to have difficulty converging to correct positions when constraints from input views are insufficient. To facilitate convergence, we introduce two geometric constraints derived from MVS outputs and recent monocular depth priors\cite{DPT}. It is difficult to directly control the geometry of Gaussians through regularizing these 3D positions. Therefore, similar to NeRF\cite{dsnerf,sparsenerf, zhu2023fsgs}, we render the depth map as a proxy by blending the z-buffer of the sorted Gaussians contributing to pixels:
\begin{equation}
\label{eq:3dgs_depth}
    d = \sum_{i\in N} d_i \alpha_i \prod_{j=1}^{i-1} (1-\alpha_j),
\end{equation}
where $d_i$ is the depth of $i-th$ Gaussian and $\alpha_i$ is identical to that in Eq. \ref{eq:3dgs_color}.
\subsubsection{Consistent Structure Regularization.}
\label{subsub_sec:consistent_structure_loss}
The MVS depths computed in Sec. \ref{subsub_sec:gaussian_init} are checked by geometric consistency and serve as the initialization for Gaussian parameters. These reliable geometric structures are expected to remain consistent throughout the optimization process. Therefore, we render the depth $D_{i}^{r}$ from Gaussian parameters with Eq. \ref{eq:3dgs_depth} for $V_{i}^{train}$ and impose a $L_{1}$ loss with MVS depths in areas with high confidence:
\begin{equation}
    \label{eq:consistent_structure_loss}
    L_{CS} = \sum |D^{r} - D^{mvs}| \odot M
\end{equation}
where $D^{mvs}$ is the MVS depth map and $M$ is the mask computed in \ref{subsub_sec:gaussian_init}. While the geometry in $D^{mvs}$ is equivalent to the initialization positions $\mathcal{X}$ and does not provide additional information about scenes, this term serves to regularize the shape to maintain consistency during optimization.

\subsubsection{Global Geometry Regularization.}
MVS depth may not be consistent in certain areas. For regions lacking consistent structure regularization, we use monocular depth priors as compensation. Specifically, we utilize the ViT-based DPT\cite{DPT}, which is trained on large-scale datasets and demonstrates strong generalization capabilities on other scenes, to predict depth map $D_{i}^{mono}$ for each view $V_{i}^{train}$. Inspired by \cite{scdepthv3,sparsenerf}, we use a random rank loss to avoid the unknown scales of monocular depths and distill the relative position relationships for whole scenes including regions without MVS depth supervision. Specifically, we random sample two disjoint batches of pixel positions $\mathcal{S}_{1}$ and $\mathcal{S}_{2}$ with the same size $n_{\mathcal{S}}$. The monocular rank loss is formulated as
\begin{equation}
    L_{mono} = \sum M'max(0,D^{r}(\mathcal{S}_{1})-D^{r}(\mathcal{S}_{2})) + \neg{M'}max(0,D^{r}(\mathcal{S}_{2})-D^{r}(\mathcal{S}_{1}))
\end{equation}
where $M' = \mathbb{I}_{D^{mono}(\mathcal{S}_{1})<D^{mono}(\mathcal{S}_{2})}$ and $n_{\mathcal{S}}$ is set to 512 in experiments.

\subsection{Training}
The total loss for optimizing 3D Gaussian parameters is
\begin{equation}
    L_{total} = L_{photo} + L_{fwd} + \beta_{1}L_{CS} + \beta_{2}L_{mono}
\end{equation}
where $\beta_{1}$ and $\beta_{2}$ are hyper-parameters. In practice, we sample unseen view every $\mathcal{E}$ iteration in which only $L_{fwd}$ is computed for optimization. Additionally, we employ the adaptive pruning and densification strategy\cite{3dgs} to dynamically adjust Gaussians during training. The training details are shown in Supp. Mat.

%% file: contents/experiment.tex
\subsection{Experimental Setup}
\noindent\textbf{Datasets.} We report results on four real-world datasets, LLFF \cite{LLFF}, DTU \cite{DTU}, NVS-RGBD\cite{sparsenerf} and T\&T (Tanks and Temples)\cite{knapitsch2017tanks,bian2023nope}. LLFF consists of 8 complex forward-facing scenes. We use every 8-th image as the held-out test set and evenly sample from the remaining views for training. We report the performance at two different resolutions of 1/8 (504 $\times$ 378) and 1/4 (1008 $\times$ 756). DTU is a multi-view dataset that contains different objects placed on a table. We use the same 15 scenes and dataset split in previous methods\cite{regnerf,sparsenerf,freenerf} and conduct experiments at resolutions of 1/4 (400 $\times$ 300) and 1/2 (800 $\times$ 600). NVS-RGBD is an RGBD dataset containing three branches, each branch consists of 8 scenes. We adopt Kinect and ZED2 for experiments with 3 given views for training and the rest for evaluation. T\&T is a complex large-scale dataset containing both outdoor and indoor scenes.  We use the first 50 frames of each scene for experiments and adhere to LLFF's dataset-split scheme.

\noindent\textbf{Metrics.} We adopt the mean of PSNR, SSIM\cite{ssim}, and LPIPS\cite{lpips} as quantitative metrics. For DTU, we follow previous few-shot methods\cite{regnerf, freenerf, sparsenerf} to remove the background with object masks in evaluation.

\noindent\textbf{Implementation Details.} Our method is built upon 3DGS\cite{3dgs} and implemented with Pytorch and CUDA. We use pre-trained MVSFormer\cite{mvsformer} to get the view-consistent depth maps of training views. MVS depths are filtered according to geometric consistency and merged into point clouds\cite{mvsnet} as the initial Gaussian positions. Considering the black background is irrelevant to objects, we use the filtered MVS depth maps with masks $M$ for DTU to only warp foreground regions. For LLFF, NVS-RGBD, and T\&T datasets, we use lama\cite{LAMA} to inpaint the warped image for border pixels. The monocular depths for input views are estimated by DPT\cite{DPT}. We report the performance of our methods with 10k (${Ours}^{\dagger}$) or 20k ($Ours$) training iterations. We densify and prune Gaussians every 100 iterations from 500 until 5k and 10k iterations, respectively. We set $\beta_{1}=0.1,\beta_{2}=0.005$ ($\beta_{2}=0.01$ for DTU) in experiments. FPS metrics in Tab. \ref{tab:all_quantitative_result} are tested on a single V100 GPU. For more details please refer to Supp. Mat.

\subsection{Comparison}
We compare performances with state-of-the-art per-scene optimization methods including RegNeRF\cite{regnerf}, FreeNeRF\cite{freenerf}, and SparseNeRF\cite{sparsenerf}. We also compare the performance of original Mip-NeRF\cite{mip_nerf} and 3DGS with sparse inputs. The training set and testing set are retained to be the same across all methods. 
\input{tables/quantitative_res_all}

\begin{figure}[h]
    \centering
    \includegraphics[width=120mm]{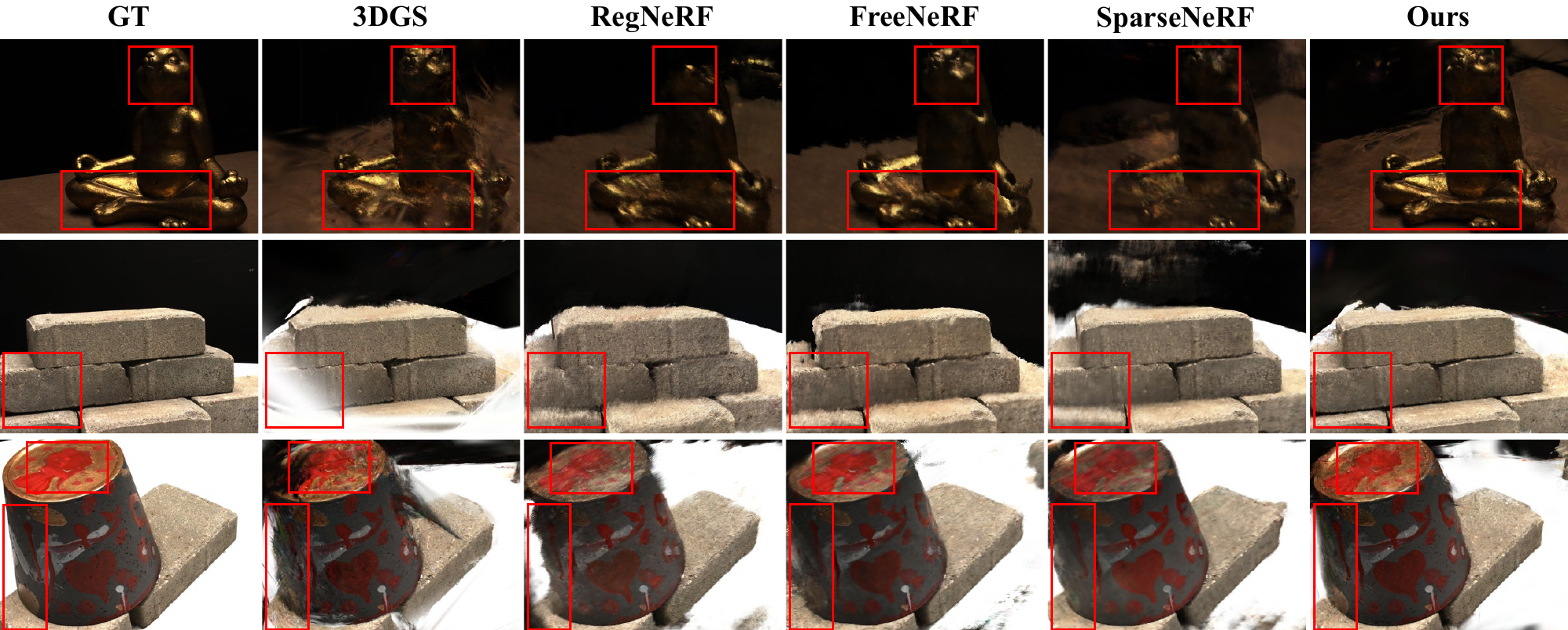}
    \caption{\textbf{Qualitative Results on DTU with 3 Input Views.}}
    \label{fig:dtu_qualitative_overall}
\end{figure}
\noindent\textbf{Comparison on DTU.} We report the quantitative results of the proposed method compared to state-of-the-art NVS methods. As shown in Tab. \ref{tab:all_quantitative_result}, our approach outperforms the compared methods in terms of rendering quality and achieves real-time rendering speed of 366 FPS (res. 1/2), which is faster than previous NeRF methods. Fig. \ref{fig:dtu_qualitative_overall} shows the rendered images in novel views under the 1/4 resolution setting. We observed that our method performs better in capturing the overall object structure and rendering finer details in local regions. Furthermore, the point-based representation enables our approach to preserve fine details even in high-resolution settings (see Fig. \ref{fig:high_res_details}). The overall quantitative results in Tab. \ref{tab:all_quantitative_result} also indicate that our approach possesses better adaptability to high-resolution rendering in sparse input scenarios. More rendering results on DTU can be found in Supp. Mat.

\begin{figure}[h]
    \centering
    \includegraphics[width=120mm]{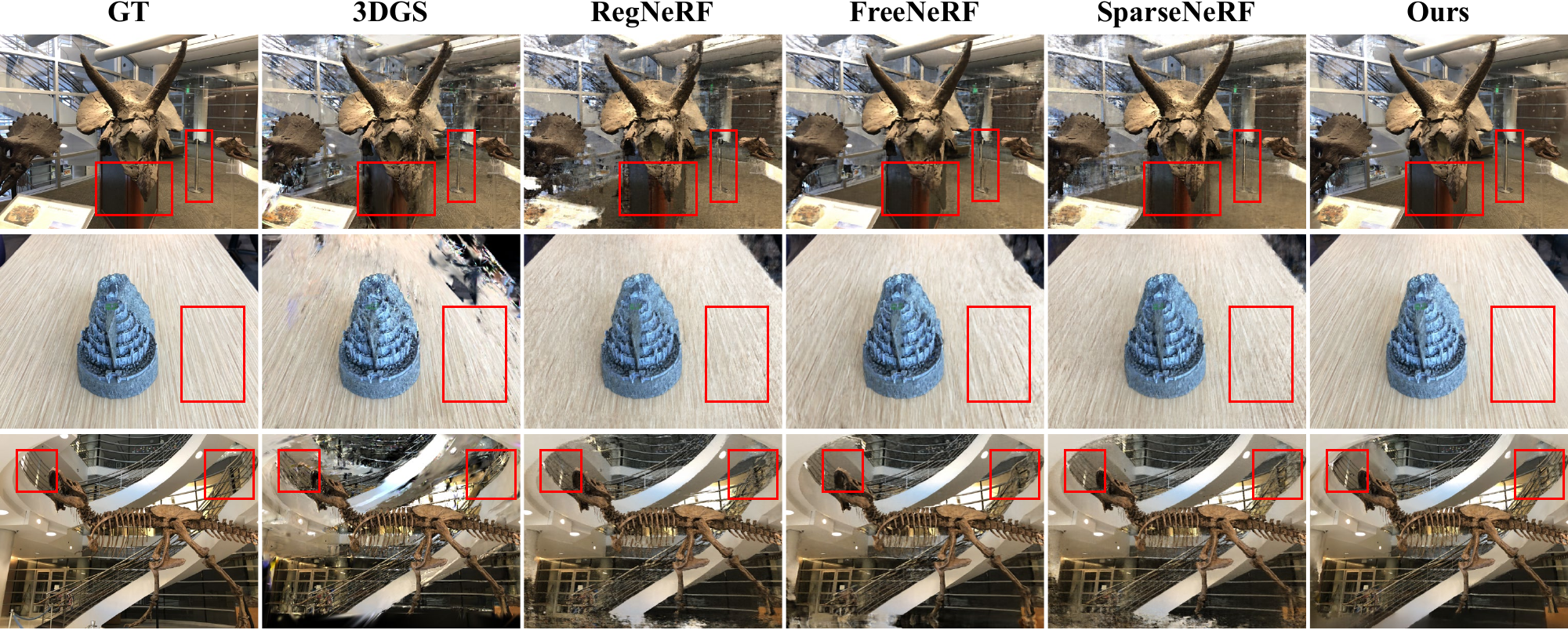}
    \caption{\textbf{Qualitative Results on LLFF with 3 Input Views.}}
    \label{fig:llff_qualitative_overall}
\end{figure}
\noindent\textbf{Comparison on LLFF.} Tab. \ref{tab:all_quantitative_result} presents the quantitative results of our method on the more complex scene-level LLFF dataset.  Our method achieves the highest scores in all rendering quality metrics across different resolutions. In particular, the higher SSIM metrics indicate that our rendering quality aligns more closely with subjective human perception. Moreover, our approach is more practical for real-world applications for it maintains a real-time rendering speed of 132 FPS (res. 1/4). Fig. \ref{fig:llff_qualitative_overall} illustrates the visual quality of novel views in LLFF under 1/8 resolution. In the horns and trex scenes, the proposed method, benefiting from constraints of multi-view consistent priors, preserves more accurate and complete structures compared to other methods. Quantitative metrics in Tab. \ref{tab:all_quantitative_result} also demonstrate our approach achieves competitive rendering quality under high resolution, particularly in high-frequency areas (see fern in Fig. \ref{fig:high_res_details}). For more visualizations on LLFF please refer to Supp. Mat.

\begin{figure}[t]
    \centering
    \includegraphics[width=120mm]{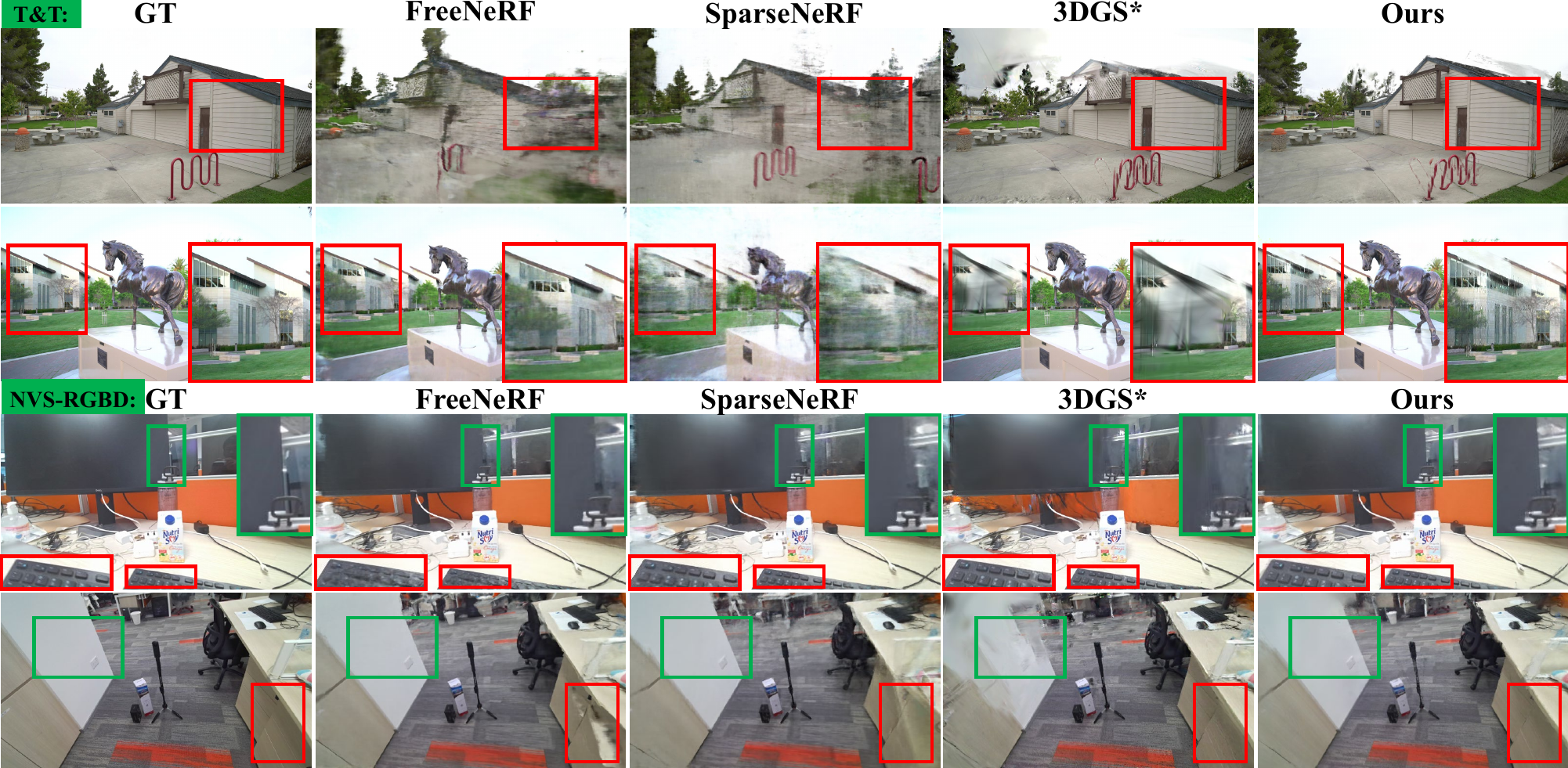}
    \caption{\textbf{Qualitative Results on T\&T and NVS-RGBD with 3 Input Views.}}
    \label{fig:tank_nvsrgbd_qualitative_overall}
\end{figure}

\input{tables/quantitative_tank_nvsrgbd}

\noindent\textbf{Comparison on T\&T and NVS-RGBD.} Results in Tab.\ref{tab:quantitative_tank_nvsrgbd} and Fig. \ref{fig:tank_nvsrgbd_qualitative_overall} show that the proposed method also achieves better performance in large-scale scenes and in-the-wild situations. Compared to 3DGS and previous NeRF-based methods, our approach adopts view-consistent constraint and is capable of rendering better details in edge areas where scene geometry changes abruptly (see visual quality on NVS-RGBD in Fig. \ref{fig:tank_nvsrgbd_qualitative_overall}). Besides, the performance on T\&T demonstrates our method's better robustness in large-scale outdoor scenes where the geometry of scenes is more complex and the variation of camera positions is quite large. Previous methods suffer from fitting the accurate geometry due to insufficient view constraints even with high-quality initialization of Gaussians (3DGS* in Tab. \ref{tab:quantitative_tank_nvsrgbd}). Our methods leverage multi-view priors for additional supervision of unseen views, which improves the quality of scene reconstruction.

\noindent\textbf{Gaussian Initialization.} In Tab. \ref{tab:all_quantitative_result} and Tab. \ref{tab:quantitative_tank_nvsrgbd}, we report quantitative results of the original 3DGS equipped with our MVS point clouds (3DGS*). 3DGS with our MVS initialization outperforms its counterpart with COLMAP points in all metrics. This shows that the learning-based MVS can provide richer initialization (please refer to Supp. Mat.), which contributes to the subsequent optimization. However, the improvement brought by only replacing better initialization is limited since the overfitting problem during few-shot optimization still exists.

\noindent\textbf{Training Views.}
Tab. \ref{tab:different_train_view_compare} shows the performance of the proposed methods in different input-view settings compared to other methods on LLFF under 1/8 resolution. Our method excels in most metrics, demonstrating robust performance across various training view selections. Moreover, the proposed method maintains competitive performance even in the 2-view extremely sparse setting.
\input{tables/different_view_num_llff}

\subsection{Ablation Studies}
In this section, we ablate our design choices on the LLFF dataset with the 3-view input setting under 1/8 resolution. 
\begin{figure}[htbp]
    \centering
    \includegraphics[width=120mm]{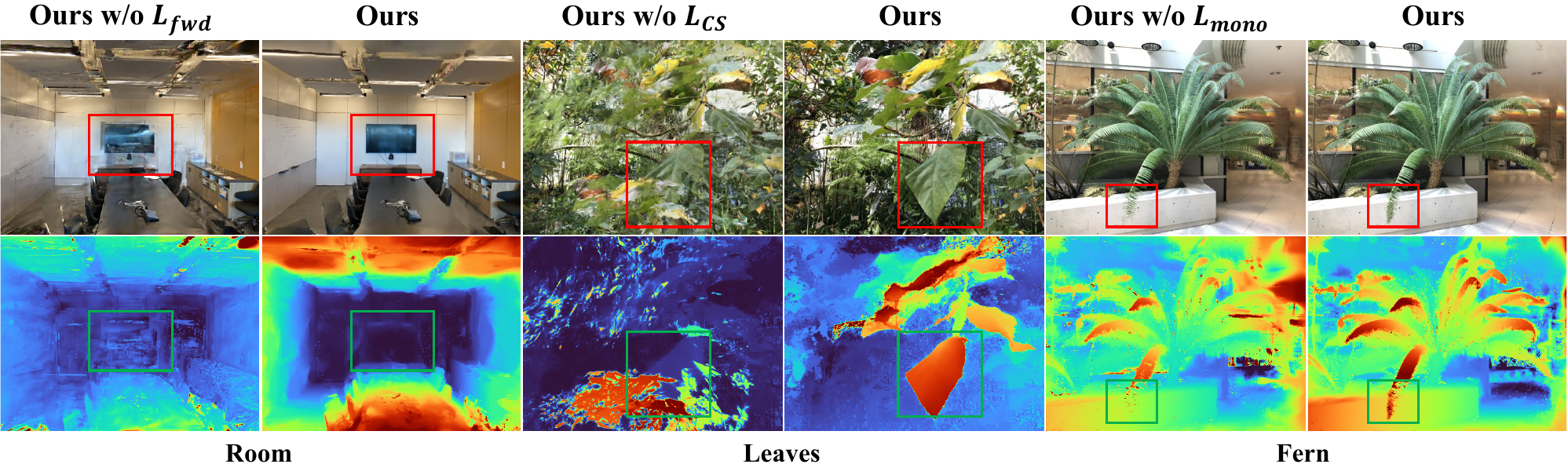}
    \caption{\textbf{Ablation Study of Visual Quality.} We compare the RGB and depth of novel views to verify the effectiveness of components in our method.}
    \label{fig:ablation_visualization}
\end{figure}

\noindent\textbf{Forward Warping Prior.} Tab. \ref{tab:ablation} shows that the performance degrades significantly when the forward-warping appearance constraint for unseen views is absent. In the room scene in Fig. \ref{fig:ablation_visualization}, the quality of the novel view appears blurry without $L_{fwd}$, indicating poor geometries of the scene caused by overfitting. Leveraging appearance priors computed from MVS, our method learns more reasonable Gaussian parameters and achieves higher-quality rendering results.

\input{tables/combined_ablation_tables}

\noindent\textbf{Consistent Structure Regularization.}
The leaves scene in Fig. \ref{fig:ablation_visualization} illustrates that the proposed consistent depth loss contributes to maintaining proper positions during optimization. Though Gaussian parameters $\mathcal{X}$ are initialized with the corresponding point clouds in the green box, they tend to overfit to improper positions in complex scenes. Our approach corrects geometric errors during optimization by utilizing confident view-consistent depths for regularization. Quantitative results in Tab. \ref{tab:ablation} also demonstrate the effectiveness of this constraint.

\noindent\textbf{Global Geometry Regularization.} Tab. \ref{tab:ablation} and Fig. \ref{fig:ablation_visualization} also demonstrates the improvement from monocular depth priors. In regions where MVS depths are filtered out due to insufficient view consistency (Sec. \ref{subsub_sec:gaussian_init}), Gaussians may encounter difficulty in fitting the geometry and appearance of scenes (as observed in the fern scene in Fig. \ref{fig:ablation_visualization}). With the compensatory constraint of monocular depth, our method can learn more reasonable geometric structures.

\noindent\textbf{Different Pre-trained MVS Models.}
We test the performance with different pre-trained models. Specifically, we test models trained on BlendedMVS\cite{blendedmvsdataset}, DTU\cite{DTU} and also trained on DTU then finetuned on BlendedMVS. Tab. \ref{tab:different_pretrained_mvs_models} demonstrates our approach is robust to the pre-trained data of MVS model.

%% file: tables/quantitative_res_all.tex
\begin{table}[t]
    \centering
    \caption{\textbf{Quantitative Results on LLFF and DTU with 3 Input Views.} Our approach achieves competitive performances across different resolutions. 3DGS* denotes 3DGS initialized with our MVS point clouds. We mark the \textbf{\colorbox[RGB]{255,179,179}{best}}, \textbf{\colorbox[RGB]{255,217,179}{second best}}, and \textbf{\colorbox[RGB]{255,255,179}{third best}} methods in cells, respectively.}
    \resizebox{\linewidth}{!}{
        \begin{tabular}{c|cccc|cccc|cccc|cccc}
        \toprule
           \multirow{3}{*}{Methods} & \multicolumn{8}{c}{LLFF} & \multicolumn{8}{c}{DTU} \\

           \cline{2-17}
           & \multicolumn{4}{c}{res. 8 $\times$} &\multicolumn{4}{c}{res. 4 $\times$} & \multicolumn{4}{|c}{res. 4 $\times$} &\multicolumn{4}{c}{res. 2 $\times$}  \\ 

             & PSNR $\uparrow$ & SSIM $\uparrow$ & LPIPS $\downarrow$ & FPS $\uparrow$ & PSNR $\uparrow$ & SSIM $\uparrow$ & LPIPS $\downarrow$ & FPS $\uparrow$ & PSNR $\uparrow$ & SSIM $\uparrow$ & LPIPS $\downarrow$ & FPS $\uparrow$ & PSNR $\uparrow$ & SSIM $\uparrow$ & LPIPS $\downarrow$ & FPS $\uparrow$ \\
            \hline
            \hline
            3DGS & 14.87 & 0.438 & 0.383 & \cellcolor{orange!25}309 & 14.40 & 0.415 & 0.428 & \cellcolor{orange!25}343 & 14.90 & 0.783 & 0.183 & \cellcolor{yellow!25}326 & 14.07 & 0.764 & \cellcolor{yellow!25}0.209 & \cellcolor{orange!25}366\\
            3DGS* & 16.16 & 0.519 & 0.331 & \cellcolor{red!25}315 & 15.92 & 0.504 & \cellcolor{orange!25}0.370 & \cellcolor{red!25}351 & 15.45 & \cellcolor{orange!25}0.814 & \cellcolor{orange!25}0.158 & \cellcolor{orange!25}359 & 15.18 & \cellcolor{orange!25}0.801 & \cellcolor{orange!25}0.177 & \cellcolor{yellow!25}355\\
            MipNeRF & 14.62 & 0.351 & 0.495 & 0.07 & 15.53 & 0.416 & 0.490 & 0.03 & 8.68 & 0.571 & 0.353 & 0.18 & 8.86 & 0.607 & 0.344 & 0.05\\
            \hline
            RegNeRF & 19.08 & 0.587 & 0.336  & 0.07 & 18.40 & 0.545 & 0.405 & 0.03 & 18.89 & 0.745 & 0.190 & 0.18 & 18.62 & 0.750 & 0.232 & 0.05\\
            FreeNeRF & \cellcolor{yellow!25}19.63 & \cellcolor{yellow!25}0.612 & \cellcolor{orange!25}0.308 & 0.07 & \cellcolor{yellow!25}19.12 & \cellcolor{orange!25}0.568 & \cellcolor{yellow!25}0.393 & 0.03 & \cellcolor{orange!25}19.92 & \cellcolor{yellow!25}0.787 & \cellcolor{yellow!25}0.182 & 0.18 & \cellcolor{orange!25}19.84 & \cellcolor{yellow!25}0.770 & 0.240 & 0.05 \\
            SparseNeRF & \cellcolor{orange!25}19.86 & \cellcolor{orange!25}0.620 & \cellcolor{yellow!25}0.329 & 0.07 & \cellcolor{orange!25}19.30 & \cellcolor{yellow!25}0.565 & 0.413 & 0.03 & \cellcolor{yellow!25}19.46 & 0.766 & 0.204 & 0.18 &  \cellcolor{yellow!25}19.63 & 0.762 & 0.242 & 0.05\\
            \hline
            \hline

            Ours$\dagger$ & \cellcolor{red!25}20.54 & \cellcolor{red!25}0.727 & \cellcolor{red!25}0.194 & \cellcolor{yellow!25}209 & \cellcolor{red!25}19.91 & \cellcolor{red!25}0.696 & \cellcolor{red!25}0.229 & \cellcolor{yellow!25}193 & \cellcolor{red!25}20.65 & \cellcolor{red!25}0.877 & \cellcolor{red!25}0.099 & \cellcolor{red!25}531 & 20.24 & \cellcolor{red!25}0.858 & \cellcolor{red!25}0.124 & \cellcolor{red!25}425 \\

            Ours & 20.39 & 0.715 & 0.203 & 132 & 19.70 & 0.681 & 0.241 & 170 & 20.50 & 0.871 & 0.106 & 365 & \cellcolor{red!25}20.26 & 0.851 & 0.130 & 366 \\
            
            \bottomrule
        \end{tabular}
}

    \label{tab:all_quantitative_result}
\end{table}

%% file: tables/quantitative_tank_nvsrgbd.tex
\begin{table}[h]
    \centering
    \caption{\textbf{Quantitative Results on T\&T and NVS-RGBD with 3 Input Views.}}
    \label{tab:quantitative_tank_nvsrgbd}
    \resizebox{0.8\linewidth}{!}{
    \begin{tabular}{c|ccc|ccc|ccc}
    \hline
    Methods    & \multicolumn{3}{c|}{NVS-RGBD(ZED2)}                                                & \multicolumn{3}{c|}{NVS-RGBD(Kinect)}                                              & \multicolumn{3}{c}{T\&T}                                    \\
               & PSNR $\uparrow$               & SSIM $\uparrow$              & \multicolumn{1}{c|}{LPIPS $\downarrow$} & PSNR $\uparrow$               & SSIM $\uparrow$              & \multicolumn{1}{c|}{LPIPS $\downarrow$} & PSNR $\uparrow$               & SSIM $\uparrow$              & LPIPS $\downarrow$           \\ 
               \hline
               \hline
    3DGS*(\textit{MVS init.})      & 21.95                         & 0.749                         & \cellcolor{orange!25}0.252            & 20.59                         & 0.770                         & 0.274                                    & \cellcolor{orange!25}22.05 & \cellcolor{orange!25}0.769 & \cellcolor{orange!25}0.207 \\
    \hline
    FreeNeRF   & \cellcolor{yellow!25}24.74 & \cellcolor{yellow!25}0.780 & 0.269                                    & \cellcolor{orange!25}26.47 & \cellcolor{orange!25}0.856 & \cellcolor{orange!25}0.206            & 18.49                         & 0.533                         & 0.463                         \\
    SparseNeRF & \cellcolor{orange!25}26.26 & \cellcolor{orange!25}0.805 & \cellcolor{yellow!25}0.261            & \cellcolor{yellow!25}26.30 & \cellcolor{yellow!25}0.848 & \cellcolor{yellow!25}0.232            & \cellcolor{yellow!25}20.62 & \cellcolor{yellow!25}0.611 & \cellcolor{yellow!25}0.414 \\ 
    \hline
    \hline
    Ours       & \cellcolor{red!25}26.62 & \cellcolor{red!25}0.841 & \cellcolor{red!25}0.185            & \cellcolor{red!25}27.04 & \cellcolor{red!25}0.887 & \cellcolor{red!25}0.151            & \cellcolor{red!25}25.57 & \cellcolor{red!25}0.846 & \cellcolor{red!25}0.139
    \\
    \bottomrule
    \end{tabular}
    }
\end{table}

%% file: tables/different_view_num_llff.tex
\begin{table}[t]
    \centering
    \caption{\textbf{Quantitative Results with Different Input Views on LLFF.} The proposed method achieves the best in most metrics under different input-view settings.}
    \resizebox{\linewidth}{!}{
        \begin{tabular}{c|ccc|ccc|ccc|ccc}
        \toprule
           \multirow{2}{*}{Methods} & \multicolumn{3}{c}{2 view} & \multicolumn{3}{c}{3 view} & \multicolumn{3}{c}{4 view} & \multicolumn{3}{c}{5 view} \\ 
             & PSNR $\uparrow$ & SSIM $\uparrow$ & LPIPS $\downarrow$ & PSNR $\uparrow$ & SSIM $\uparrow$ & LPIPS $\downarrow$ & PSNR $\uparrow$ & SSIM $\uparrow$ & LPIPS $\downarrow$ & PSNR $\uparrow$ & SSIM $\uparrow$ & LPIPS $\downarrow$ \\
            \hline
            \hline
            3D-GS & 13.39 & 0.322 & 0.459 & 14.87 & 0.438 & 0.383 & 16.78 & 0.541 & 0.321 & 18.21 & 0.606 & 0.284\\
            Mip-NeRF & 12.44 & 0.205 & 0.581 & 14.62 & 0.351 & 0.495 & 19.41 & 0.628 & 0.311 & 20.84 & 0.667 & 0.284\\
            \hline
            RegNeRF & 16.32 & 0.404 & 0.442 & 19.08 & 0.587 & 0.336 & 21.10 & 0.692 & \cellcolor{yellow!25}0.271 & 21.81 & \cellcolor{yellow!25}0.712 & \cellcolor{yellow!25}0.258\\
            FreeNeRF & \cellcolor{yellow!25}16.99 & \cellcolor{yellow!25}0.473 & \cellcolor{orange!25}0.366 & \cellcolor{yellow!25}19.63 & \cellcolor{yellow!25}0.612 & \cellcolor{orange!25}0.308 & \cellcolor{orange!25}21.28 & \cellcolor{orange!25}0.703 & \cellcolor{orange!25}0.255 & \cellcolor{red!25}22.57 & \cellcolor{orange!25}0.725 & \cellcolor{orange!25}0.247\\
            SparseNeRF & \cellcolor{orange!25}17.44 & \cellcolor{orange!25}0.447 & \cellcolor{yellow!25}0.423 & \cellcolor{orange!25}19.86 & \cellcolor{orange!25}0.620 & \cellcolor{yellow!25}0.329 & \cellcolor{red!25}21.42 & \cellcolor{yellow!25}0.696 & 0.283 & \cellcolor{yellow!25}21.89 & 0.695 & 0.288\\
            \hline
            \hline
            Ours & \cellcolor{red!25}18.53 & \cellcolor{red!25}0.607 & \cellcolor{red!25}0.280 & \cellcolor{red!25}20.39 & \cellcolor{red!25}0.715 & \cellcolor{red!25}0.203 & \cellcolor{orange!25}21.28 & \cellcolor{red!25}0.750 & \cellcolor{red!25}0.180 & \cellcolor{orange!25}22.18 & \cellcolor{red!25}0.773 & \cellcolor{red!25}0.164\\
            \bottomrule
        \end{tabular}
    }
    \label{tab:different_train_view_compare}
\end{table}

%% file: tables/combined_ablation_tables.tex
\begin{table}[t]

\begin{minipage}[t]{0.45\linewidth}
    \centering
    
    \caption{\textbf{Quantitative Ablation Results on the LLFF Dataset.}}
    \resizebox{0.73 \linewidth}{!}{
    \begin{tabular}{c | c c c}
    \toprule
    Methods & PSNR $\uparrow$   & SSIM $\uparrow$   & LPIPS $\downarrow$ \\ \hline
    Baseline    & 14.87 & 0.438 & 0.383   \\ 
    Baseline w/ $mvs$ $pc$    & 16.16 & 0.519 & 0.331   \\
    \hline
    Ours w/o $L_{fwd}$    & 17.32 & 0.563 & 0.306   \\ 
    Ours w/o $L_{CS}$    & 20.06 & 0.699 & 0.218   \\ 
    Ours w/o $L_{mono}$    & 20.17 & 0.708 & 0.209   \\ 
    \hline
    Ours    & \textbf{20.39} & \textbf{0.715} & \textbf{0.203}   \\ 
    \bottomrule
    \end{tabular}
    }
    \label{tab:ablation}
\end{minipage}
\begin{minipage}[t]{0.5\linewidth}
    \centering
    
    \caption{\textbf{Impact of pre-trained MVS models on LLFF Dataset.}}
    \resizebox{1.0 \linewidth}{!}{
    \begin{tabular}{c | c c c}
    \toprule
    LLFF & PSNR $\uparrow$   & SSIM $\uparrow$   & LPIPS $\downarrow$ \\ 
    \hline
    Baseline    & 14.87 & 0.438 & 0.383   \\ 
    \hline
    DTU    & \textbf{20.39} & 0.715 & 0.203   \\ 
    BlendedMVS    & 20.17 & 0.709 & 0.213   \\ 
    DTU and BlendedMVS ft    & 20.34 & \textbf{0.722} & \textbf{0.202}   \\ 
    \bottomrule
    \end{tabular}
    }
    \label{tab:different_pretrained_mvs_models}
\end{minipage}

\end{table}

%% file: contents/conclusion.tex
In this paper, we propose a novel paradigm for real-time few-shot novel view synthesis based on 3D Gaussian Splatting. The key insight is to excavate the multi-view cues derived from the data-driven MVS method for limited input views. The geometry computed from MVS serves as the initialization of Gaussians for capturing more information about scenes. To alleviate overfitting, we introduce a forward warping method to generate appearance priors for unseen views. To facilitate convergence during optimization, we propose a view-consistent depth loss in regions with confident geometry and also a monocular depth loss for additional geometry constraints. Experiments show that our method achieves state-of-the-art performance in the task of few-shot novel view synthesis.

\noindent\textbf{Limitation.} The proposed method mitigates the overfitting problem during optimization with the view-consistent appearance priors, but it does not deal with the light variation of non-Lambertian surfaces.

%% file: contents/acknowledgments.tex
\section*{Acknowledgments}
This work is financially supported by Outstanding Talents Training Fund in Shenzhen, Shenzhen Science and Technology Program-Shenzhen Cultivation of Excellent Scientific and Technological Innovation Talents project (Grant No. RCJC20200714114435057), Shenzhen Science and Technology Program-Shenzhen Hong Kong joint funding project (Grant No. SGDX20211123144400001), Guangdong Provincial Key Laboratory of Ultra High Definition Immersive Media Technology, National Natural Science Foundation of China U21B2012, R24115SG MIGU-PKU META VISION TECHNOLOGY INNOVATION LAB.
Jianbo Jiao is supported by the Royal Society Short Industry Fellowship (SIF$\setminus$R1$\setminus$231009).

%% file: contents/supplementary.tex
\section{More Results}
\input{supp_contents/additional_results}

\section{Experiment Details}
\input{supp_contents/exp_details}

%% file: supp_contents/additional_results.tex
\subsection{Visual Quality}
\begin{figure}[h]
    \centering
    \includegraphics[width=120mm]{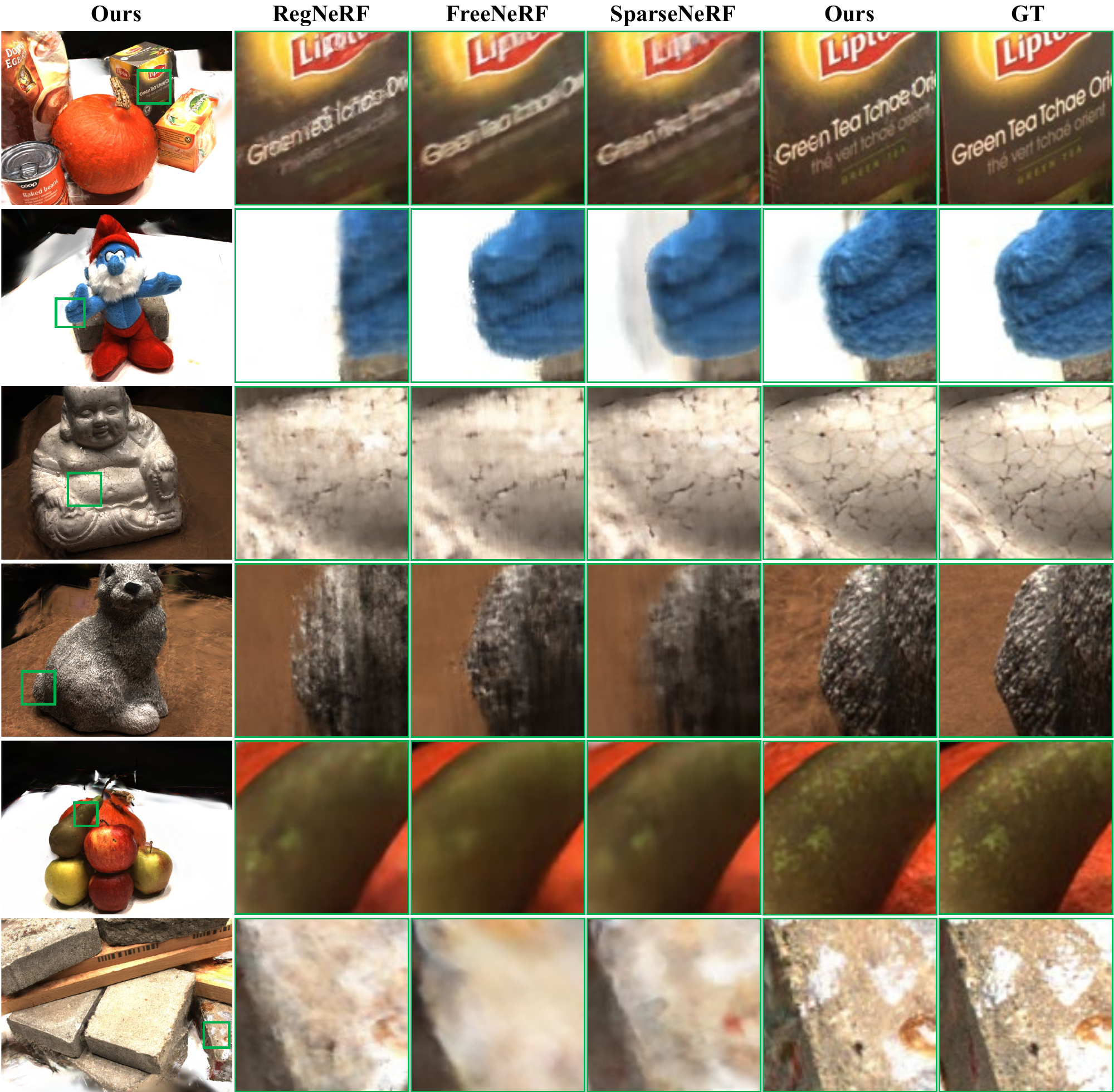}
    \caption{\textbf{Qualitative Results on DTU with 3 Input Views (res. 1/2).}}
    \label{supp_fig:dtu3_2x_details}
\end{figure}

\begin{figure}[h]
    \centering
    \includegraphics[width=120mm]{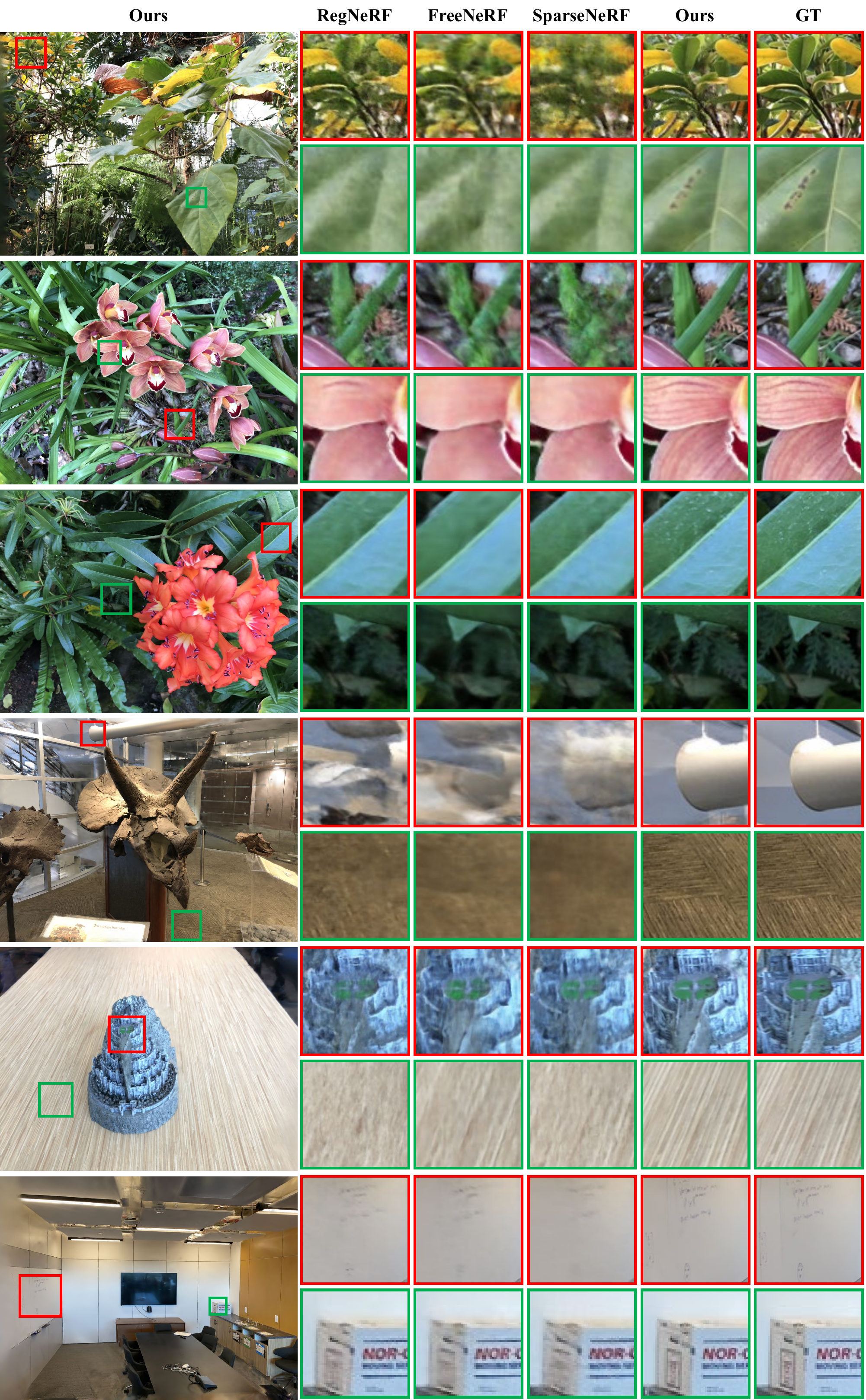}
    \caption{\textbf{Qualitative Results on LLFF with 3 Input Views (res. 1/4).}}
    \label{supp_fig:llff3_4x_details}
\end{figure}

\begin{figure}[t]
    \centering
    \includegraphics[width=120mm]{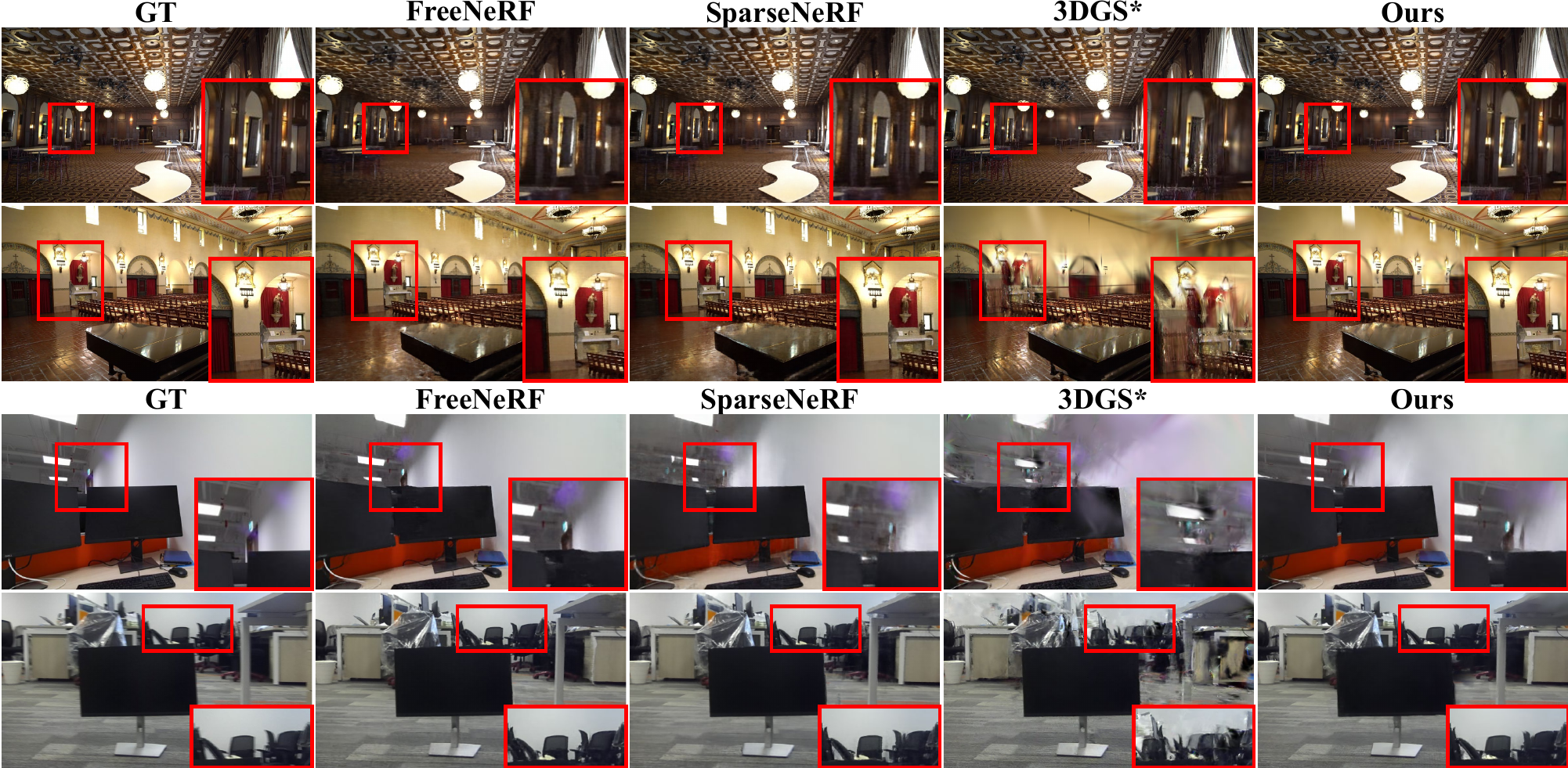}
    \caption{\textbf{Qualitative Results on T\&T (the first two rows) and NVS-RGBD (the last two rows) with 3 Input Views.}}
    \label{supp_fig:tank3_nvsrgbd3}
\end{figure}

Fig. \ref{supp_fig:dtu3_2x_details}, Fig. \ref{supp_fig:llff3_4x_details} illustrate the qualitative results of our approach on DTU and LLFF datasets under high-resolution settings with three input views. These visualizations demonstrate our method's ability to capture finer details in high-frequency regions compared to NeRF-based approaches. Fig. \ref{supp_fig:tank3_nvsrgbd3} shows more rendering results on T\&T and NVS-RGBD datasets, demonstrating our method's robustness in complex large-scale scenes and in-the-wild scenarios.

\clearpage

\subsection{Different Input Views} 
\begin{figure}[h]
    \centering
    \includegraphics[width=120mm]{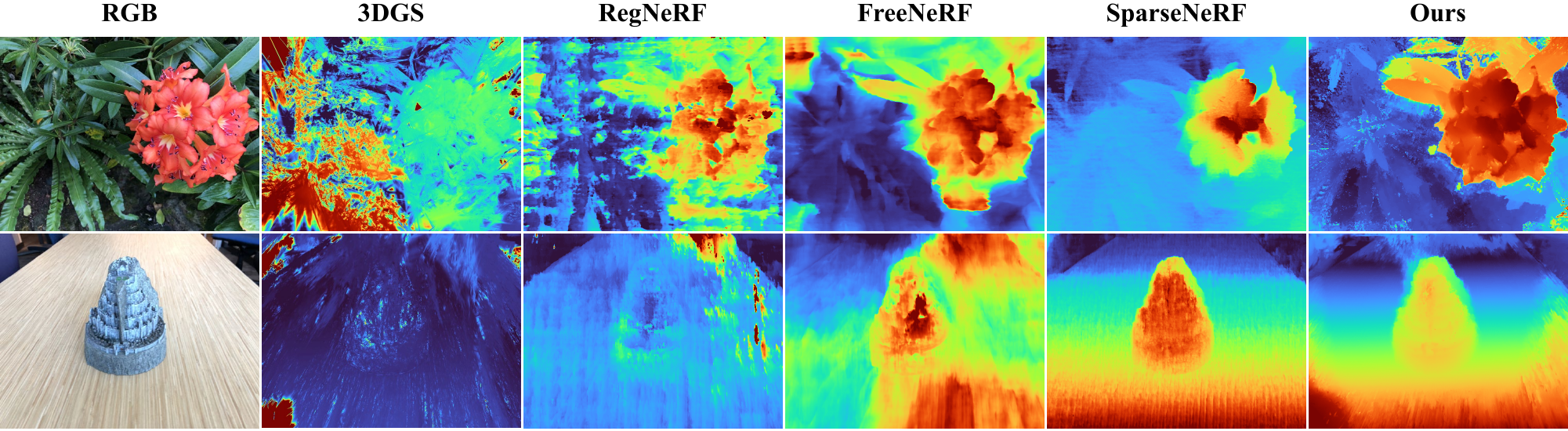}
    \caption{\textbf{Depths Visualization on LLFF with 2 Input Views (res. 1/8).}}
    \label{supp_fig:llff2_8x_depth_vis}
\end{figure}

\begin{figure}[htbp]
    \centering
    \includegraphics[width=120mm]{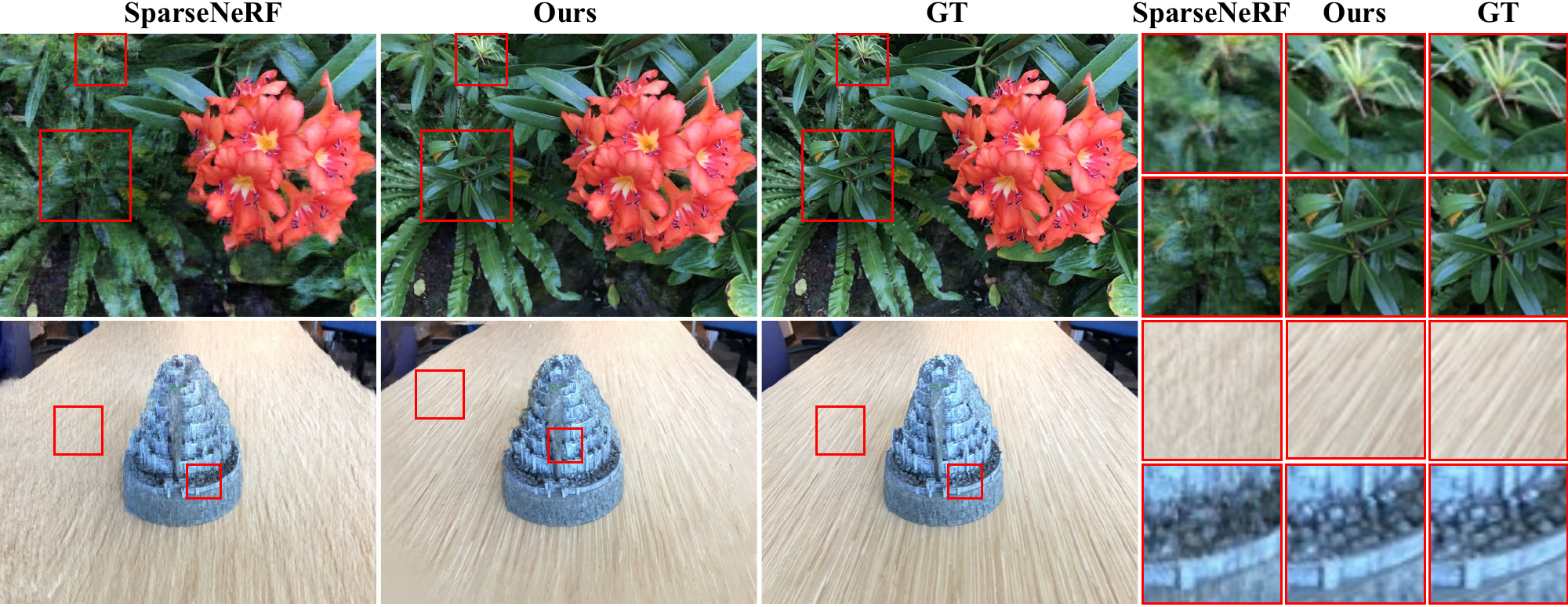}
    \caption{\textbf{Qualitative Results on LLFF with 2 Input Views (res. 1/8).}}
    \label{supp_fig:llff2_8x_overall}
\end{figure}

\begin{figure}[htbp]
    \centering
    \includegraphics[width=120mm]{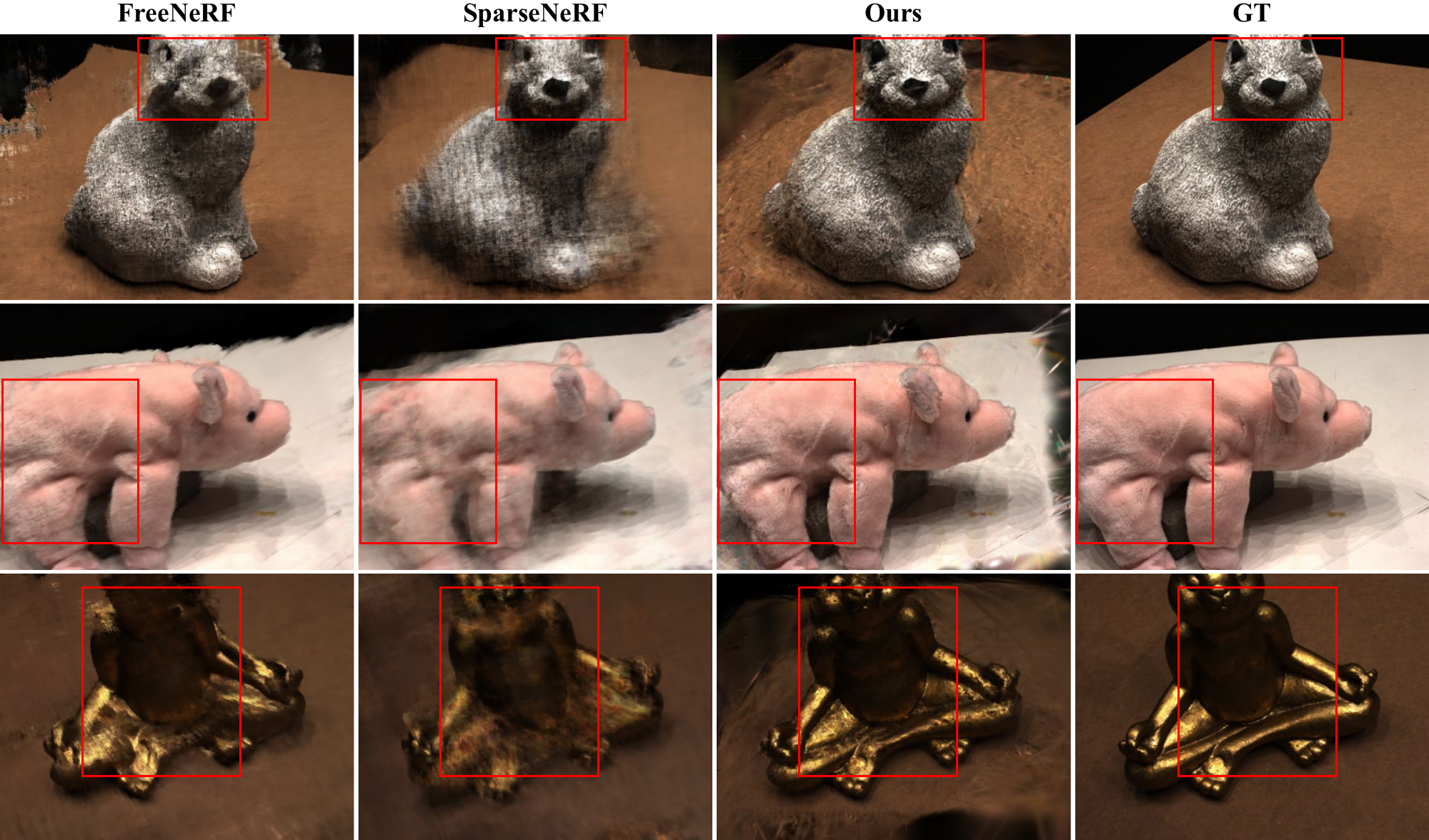}
    \caption{\textbf{Qualitative Results on DTU with 2 Input Views (res. 1/4).}}
    \label{supp_fig:dtu2_4x_overall}
\end{figure}

\begin{figure}[htbp]
    \centering
    \includegraphics[width=120mm]{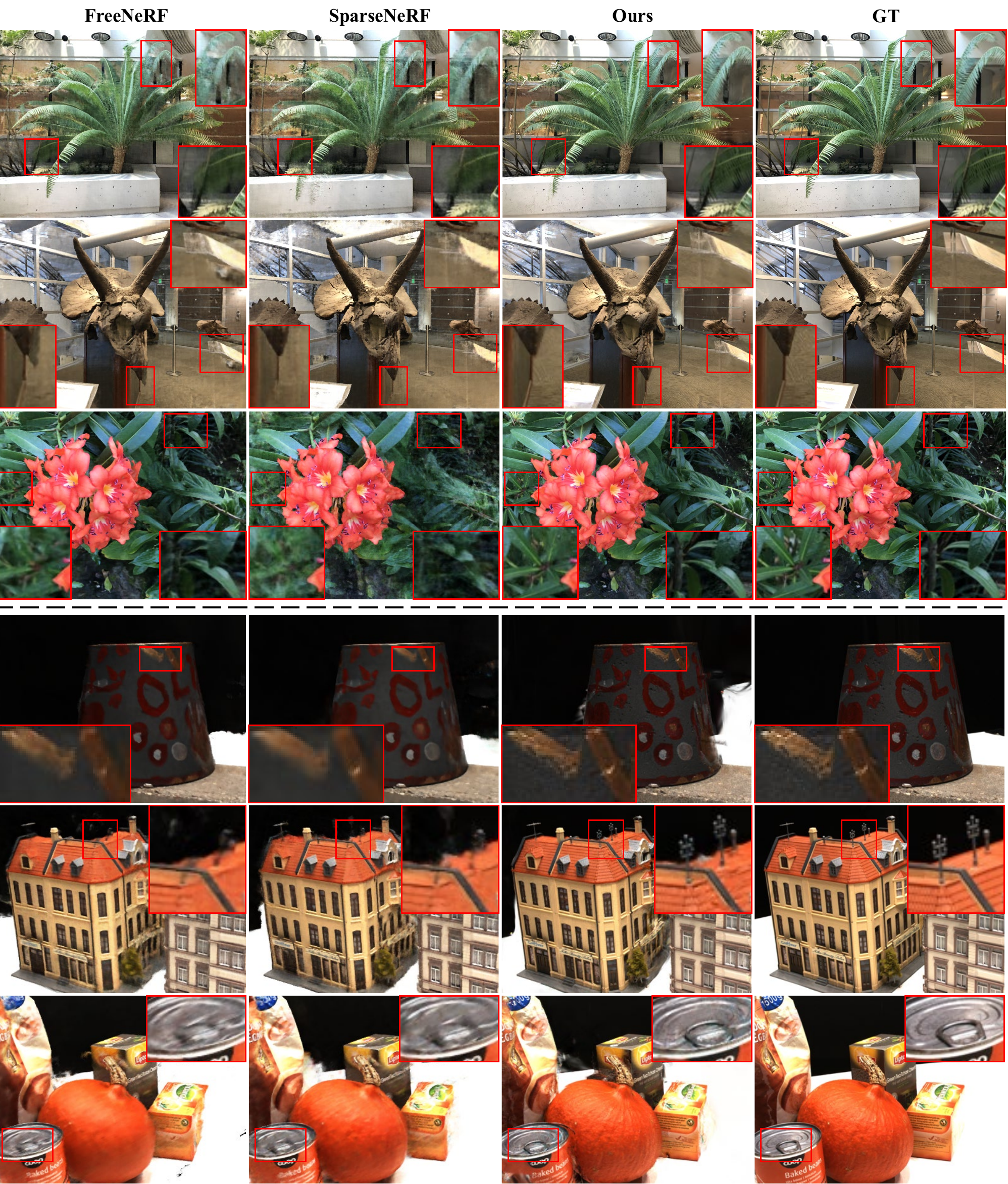}
    \caption{\textbf{Qualitative Results on LLFF (res. 1/8, the first three rows) and DTU (res. 1/4, the last three rows) with 5 Input Views.}}
    \label{supp_fig:llff5_8x_dtu5_4x_overall}
\end{figure}

Tab. \ref{supp_tab:different_train_view_compare_dtu} shows the quantitative results with different input views on DTU (res. 1/4). Similar to the results in Tab. \ref{tab:different_train_view_compare}, our method achieves competitive results in most metrics on DTU, which demonstrates our robustness in different training views. Fig. \ref{supp_fig:llff2_8x_overall} and Fig. \ref{supp_fig:dtu2_4x_overall} illustrate the visual quality with only 2-view inputs. Our method can still render reasonable results, even in such an extremely sparse situation. The visualizations of rendered depths in novel views (Fig. \ref{supp_fig:llff2_8x_depth_vis}) suggest that our approach can catch accurate geometric structures of scenes in few-shot situations. Fig. \ref{supp_fig:llff5_8x_dtu5_4x_overall} gives the qualitative results with more input views (5 views). It shows that our method still outperforms in finer details regions.

\input{supp_contents/supp_tables/different_view_num_dtu}

\clearpage

\subsection{Visualization of Different Gaussian Initialization} 
\begin{figure}[h]
    \centering
    \includegraphics[width=120mm]{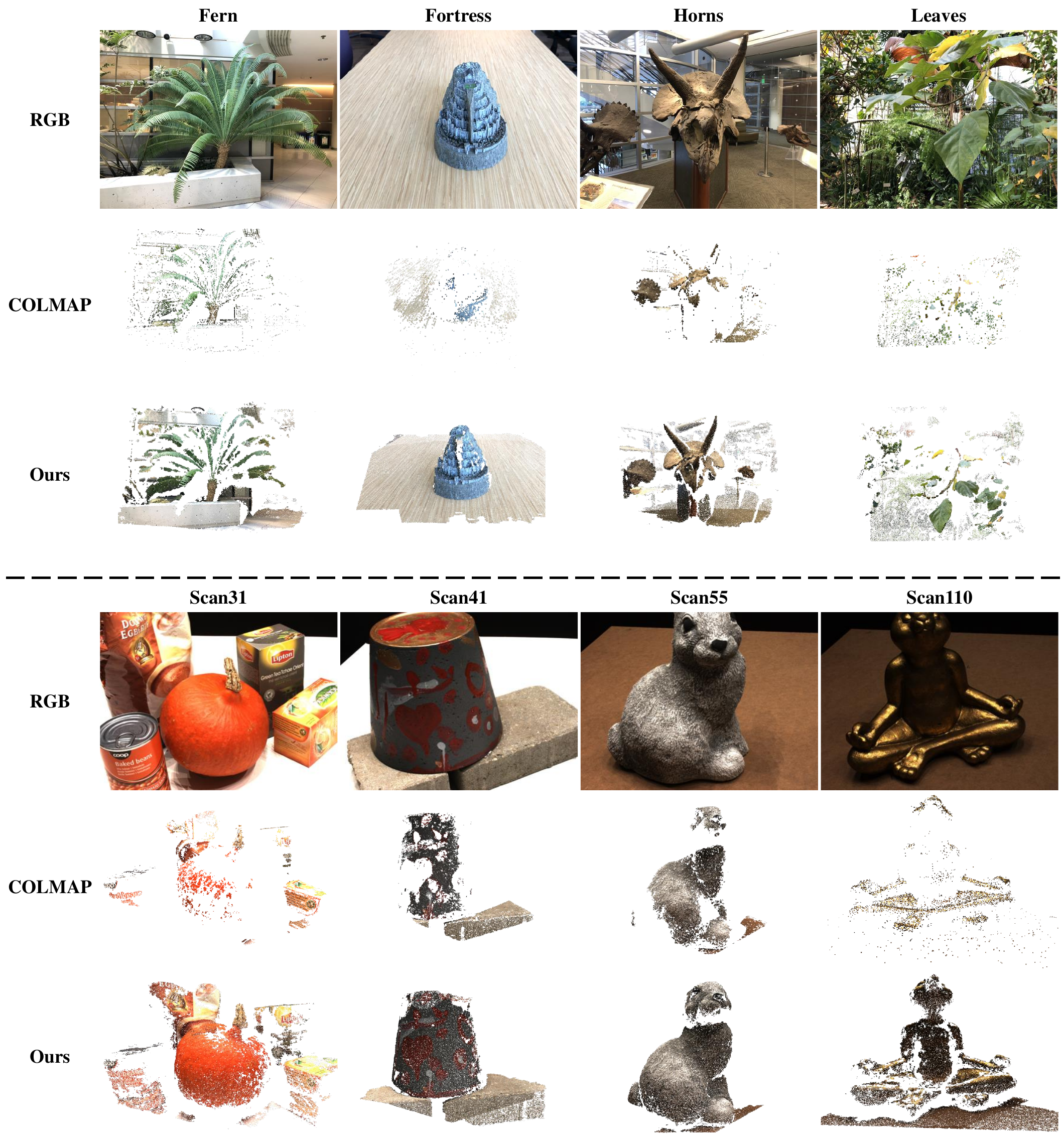}
    \caption{\textbf{Visualizations of Initial Points for LLFF (the first three rows) and DTU (the last three rows) with 3 Input Views.}}
    \label{supp_fig:llff3_dtu3_init_pc}
\end{figure}

Fig. \ref{supp_fig:llff3_dtu3_init_pc} compares visualizations of COLMAP points and our points derived from MVSFormer\cite{mvsformer} on LLFF and DTU datasets. It shows that learning-based methods can offer more comprehensive initial positions in few-shot scenarios, thereby facilitating subsequent optimization of Gaussian parameters, as indicated by the metrics in Tab. \ref{tab:all_quantitative_result}.

\clearpage

%% file: supp_contents/supp_tables/different_view_num_dtu.tex
\begin{table}[h]
    \centering
    \caption{\textbf{Quantitative Results with Different Input Views on DTU.}}
    \resizebox{\linewidth}{!}{
        \begin{tabular}{c|ccc|ccc|ccc|ccc}
        \toprule
           \multirow{2}{*}{Methods} & \multicolumn{3}{c}{2 view} & \multicolumn{3}{c}{3 view} & \multicolumn{3}{c}{4 view} & \multicolumn{3}{c}{5 view} \\ 
             & PSNR $\uparrow$ & SSIM $\uparrow$ & LPIPS $\downarrow$ & PSNR $\uparrow$ & SSIM $\uparrow$ & LPIPS $\downarrow$ & PSNR $\uparrow$ & SSIM $\uparrow$ & LPIPS $\downarrow$ & PSNR $\uparrow$ & SSIM $\uparrow$ & LPIPS $\downarrow$ \\
            \hline
            \hline
            3DGS & 12.91 & \cellcolor{orange!25}0.728 & \cellcolor{orange!25}0.226 & 14.90 & \cellcolor{yellow!25}0.783 & \cellcolor{yellow!25}0.183 & 18.01 & \cellcolor{orange!25}0.840 & \cellcolor{orange!25}0.132 & 21.62 & \cellcolor{orange!25}0.886 & \cellcolor{orange!25}0.092\\
            Mip-NeRF & 8.39 & 0.561 & 0.365 & 8.68 & 0.571 & 0.353 & 15.57 & 0.692 & 0.245 & 18.35 & 0.758 & 0.202\\
            \hline
            RegNeRF & \cellcolor{yellow!25}16.51 & 0.711 & \cellcolor{yellow!25}0.236 & 18.89 & 0.745 & 0.190 & 19.89 & 0.783 & \cellcolor{yellow!25}0.177 & 21.91 & \cellcolor{yellow!25}0.829 & \cellcolor{yellow!25}0.147\\
            FreeNeRF & 15.47 & 0.694 & 0.247 & \cellcolor{orange!25}19.92 & \cellcolor{orange!25}0.787 & \cellcolor{orange!25}0.182 & \cellcolor{orange!25}21.04 & \cellcolor{yellow!25}0.802 & 0.185 & \cellcolor{orange!25}22.35 & 0.819 & 0.175\\
            SparseNeRF & \cellcolor{orange!25}17.21 & \cellcolor{yellow!25}0.715 & 0.238 & \cellcolor{yellow!25}19.46 & 0.766 & 0.204 & \cellcolor{yellow!25}20.69 & 0.800 & 0.190 & \cellcolor{yellow!25}22.05 & 0.820 & 0.178\\
            \hline
            \hline
            Ours & \cellcolor{red!25}17.55 & \cellcolor{red!25}0.823 & \cellcolor{red!25}0.147 & \cellcolor{red!25}20.50 & \cellcolor{red!25}0.871 & \cellcolor{red!25}0.106 & \cellcolor{red!25}21.57 & \cellcolor{red!25}0.887 & \cellcolor{red!25}0.091 & \cellcolor{red!25}22.87 & \cellcolor{red!25}0.899 & \cellcolor{red!25}0.080\\
            \bottomrule
        \end{tabular}
    }
    \label{supp_tab:different_train_view_compare_dtu}
\end{table}

%% file: supp_contents/exp_details.tex
\subsection{Datasets} 
\label{supp_datasets}
We follow previous methods\cite{regnerf,freenerf,sparsenerf} for the splits of LLFF and DTU datasets. For LLFF, we use every 8-th image as the held-out test set, and evenly sample from the remaining views for training. For DTU, we adhere to \cite{regnerf,freenerf,sparsenerf} to evaluate the proposed method on 15 selected scenes. The scan IDs include 8, 21, 30, 31, 34, 38, 40, 41, 45, 55, 63, 82, 103, 110, and 114. The images with IDs (all image IDs start from 0): 25, 22, 28, 40, 44, 48, 0, 8, 13 are used for training. We select the first 2,3,4,5 image IDs as input views in our experiments. Images with IDs: 1, 2, 9, 10, 11, 12, 14, 15, 23, 24, 26, 27, 29, 30, 31, 32, 33, 34, 35, 41, 42, 43, 45, 46, 47 are used as the novel views in evaluation. For NVS-RGBD, the 3 training views are fixed and we use images with the resolution of 960$\times$540 and 640$\times$360 for ZED2 and Kinect branches, respectively. For T\&T, we conduct experiments on images with a resolution of 960$\times$540.

\subsection{Metrics}
We use the formula $-10\cdot\log_{10}(MSE)$ to compute the metric PSNR. We uniformly adopt the scikit-image\footnote{https://scikit-image.org/} python package to compute the SSIM metrics for comparison. We uniformly use lpips package\footnote{https://pypi.org/project/lpips/} with VGG model to compute LPIPS metrics in experiments.

\input{tables/algorithm}
\subsection{Trainging Details}
Alg. \ref{alg:training_process} illustrates the details of the training procedure. We set the unseen-view sampling interval $\mathcal{E}=3$ in our experiments. We follow the setting in SparseNeRF\cite{sparsenerf} to predict monocular depth maps by DPT Hybrid model.

\subsection{Details for MVS}
\subsubsection{Basic Settings} 
We use the official codebase\footnote{https://github.com/ewrfcas/MVSFormer} of MVSFormer\cite{mvsformer} to estimate each view's depth $D^{mvs}$ with default settings. The input images are resized to the default resolution 1152 $\times$ 1536 for MVSFormer. We adopt scene bounds provided by datasets as depth ranges. The number of depth hypothesis planes is set to the default 192 for all datasets.

\subsubsection{Pretrained Models} 
We use official MVSFormer model pre-trained on DTU for our experiments on LLFF. For NVS-RGBD and T\&T, we adopt the model pre-trained on DTU and fine-tuned on BlendedMVS. For our experiments on DTU, we exclude DTU scenes from the training data of MVSFormer that overlap with the selected scenes mentioned in Sec. \ref{supp_datasets} and then fine-tuned on BlendedMVS. 

\subsubsection{Geometric Filter} 
We utilize a geometric filter\cite{mvsnet} to identify regions exhibiting high view consistency, followed by reprojecting the estimated depth $D^{mvs}$ to highlight these areas. Specifically, we project a reference view pixel $p_{1}$ based on its depth $d_{1}$ to pixel $p_{i}$ in a source view $V_{i}$. Then we project $p_{i}$ back to the reference view by $p_{i}$'s depth $d_{i}$ in source view's $D_{i}^{mvs}$ to get the reprojected coordinate $p_{reproj}$ and reprojected depth $d_{reproj}$. We consider depth estimation $d_{1}$ of $p_{1}$ exhibits high-consistency with source view $V_{i}$, when it satisfies $|p_{reproj}-p_{1}|<1$ and $|d_{reproj}-d_{1}|/d_{1}<0.01$\cite{mvsnet}. For the masks $\{M_{i}\}$, we require the predicted depth to be highly consistent with at least two known viewpoints.

For the fusion of depth maps, We use the method mentioned in MVSNet\cite{mvsnet} to merge depths into point clouds for the initialization of 3DGS. We down-sample the initial point cloud with a downsampling rate of 0.1 for memory conservation.

\subsection{Computational Time Costs}
We test the computational time costs of our method on a single V100 GPU. The MVS depth estimation and filtering processes take approximately 0.57 seconds and 0.19 seconds per image for LLFF (res. 1/8) and DTU (res. 1/4), respectively. The time costs for depth maps fusion (3 views) are 2.9 seconds and 3.2 seconds per scene for LLFF (res. 1/8) and DTU (res. 1/4) datasets, respectively. Comparatively, the original COLMAP takes around 293 seconds and 83 seconds per scene for dense point cloud reconstruction on the LLFF and DTU datasets.

%% file: tables/algorithm.tex
\begin{algorithm}[h]
\SetAlgoLined
\KwData{$N$ input views $\{V_{i}^{train}\}_{i=1}^{i=N}$, ground-truth images $\{I_{i}\}_{i=1}^{i=N}$, \\camera intrinsics $K$, camera poses $\{P_{i}\}_{i=1}^{i=N}$}
\KwResult{Optimized Gaussian parameters $\mathcal{G}$}
Estimate view-consistent depths $\{D_{i}^{mvs}\}_{i=1}^{i=N}$ for $\{V_{i}^{train}\}_{i=1}^{i=N}$ by MVSFormer\;
Compute high consistency masks $\{M_{i}\}_{i=1}^{i=N}$ based on reprojected errors of $\{D_{i}^{mvs}\}_{i=1}^{i=N}$, estimate monocular depths $\{D_{i}^{mono}\}_{i=1}^{i=N}$ by DPT \;
Merge masked depths into point clouds $\mathcal{P}$ and initialize $\mathcal{G}$ with $\mathcal{P}$ \;
Sample unseen views $\{V_{i}^{unseen}\}_{i=1}^{i=M}$ randomly according to $\{V_{i}^{train}\}_{i=1}^{i=N}$ \;

\For{it from 1 to num\_iters}
{
    \eIf{$\text{it}~\%~\mathcal{E} \neq 0$}
    {
        Choose a $V_{i}^{train}$ randomly \;
        Render image $\hat{I_{i}}$ and depth $\hat{D_{i}}$ for $V_{i}^{train}$ based on $\mathcal{G}$ by splatting \;
        $\mathcal{L} \gets L_{photo}(\hat{I_{i}},I_{i})$ \;
        $\mathcal{L} \gets \mathcal{L} + L_{CS}(\hat{D_{i}},D^{mvs}_{i},M_{i})$ \;
        $\mathcal{L} \gets \mathcal{L} + L_{mono}(\hat{D_{i}},D^{mono}_{i})$ \;
    }
    {
        Choose a $V_{i}^{unseen}$ randomly \;
        Choose a $V_{j}^{train}$ randomly, compute appearance prior by $I'_{i} = fwd(I_{j}^{train}, D_{j}^{mvs}, P_{j}^{train}, P_{i}^{unseen})$\;
        Render image $\hat{I_{i}}$ for $V_{i}^{unseen}$ based on $\mathcal{G}$ by splatting \;
        $\mathcal{L} \gets L_{fwd}(\hat{I_{i}},I'_{i})$ \;
    }
    $\nabla_\mathcal{G} \mathcal{L} \gets \frac{\partial \mathcal{L}}{\partial \mathcal{G}}$ \;
    \If{IsRefinementIteration(it)}
    {
        PruneAndDensify($\mathcal{G},\nabla_\mathcal{G} \mathcal{L}$) \;
    }
    Update parameters: $\mathcal{G} \gets Adam(\mathcal{G}, \nabla_\mathcal{G} \mathcal{L})$\;
}
\caption{Optimizing on a single scene}
\label{alg:training_process}
\end{algorithm}